\pdfoutput=1

\documentclass[11pt]{article}

\usepackage[final]{acl}

\usepackage{times}
\usepackage{latexsym}

\usepackage[T1]{fontenc}

\usepackage[utf8]{inputenc}

\usepackage{microtype}

\usepackage{inconsolata}

\usepackage{multirow}
\usepackage{colortbl}

\usepackage{xspace}

\usepackage{amsmath}
\usepackage{amsfonts}
\usepackage{graphicx,subfigure}
\usepackage{ctable}
\usepackage{comment}
\usepackage{multicol}
\usepackage{epsfig}
\usepackage{pifont}
\usepackage{dsfont}
\usepackage{makecell}
\usepackage{adjustbox}
\usepackage{txfonts}

\usepackage{float}

\usepackage[boxed, ruled, vlined, linesnumbered]{algorithm2e}
\usepackage{amstext}
\usepackage{xcolor}
\usepackage{pict2e}

\usepackage{amssymb}

\title{Consistent yet Contextual: Learning Author-Level Mixtures of Style for Authorship Attribution}

\author{
  Omkar Yadav \\
  University of Michigan \\
  \texttt{oyadav@umich.edu} \And
  Michael Jiang \\
  University of Michigan \\
  \texttt{mjjiang@umich.edu} \And
  Kenan Alkiek \\
  University of Michigan \\
  \texttt{kalkiek@umich.edu} \AND
  Jian Zhu \\
  University of British Columbia \\
  \texttt{jian.zhu@ubc.ca} \And
  David Jurgens \\
  University of Michigan \\
  \texttt{jurgens@umich.edu}
}
\begin{document}
\maketitle

\begin{abstract}

Approaches to authorship verification or attribution typically model the style of a document, asking whether two documents were likely written by the same author. While recent approaches have been performant, such approaches largely ignore modeling the authors themselves and how varied their style is. Here, we introduce multiple alternatives for modeling the author and demonstrate that allowing an author's representation to capture multiple consistent but distinct writing styles yields significant performance improvements. Using our new authorship model, we then examine \textit{when} an author writes in a way that is most idiomatic of their style across four scenarios: (i) variation in the size of their writing audiences, (ii) temporal distance from when the model was trained, (iii) the effect of preceding context, and (iv) the effect of genre norms on style. Across all four experiments, we demonstrate that authors vary substantially in how idiolectic their writing is. Our results have clear implications for both the modeling and data used for authorship verification and attribution, showing that authors need to be represented with their multiple styles and that some settings produce text that is much more informative about who wrote that text.

\end{abstract}

\section{Introduction}
\label{sec:introduction}

\begin{figure}
    \centering
    \includegraphics[width=0.9\linewidth]{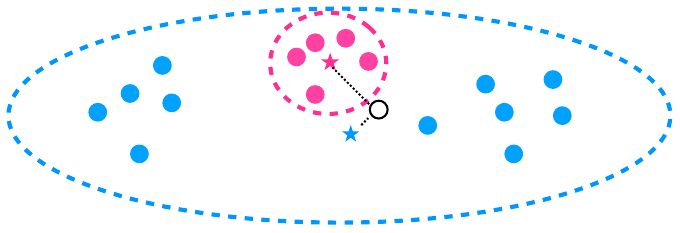}
    \caption{In authorship attribution, documents (above: circles) are embedded as style vectors. Most approaches attribute a new document (the empty circle) to the nearest neighbor or an author's centroid (stars). However, these methods overlook that authors often employ multiple styles. For instance, the new document here is nearest the blue author, even though fuchsia is the likely true source. Here, we develop models designed to mitigate this stylistic complexity. }
    \label{fig:intro-fig}
\end{figure}

Many authors have a distinct writing style that is unique to them. 
Models for authorship attribution or verification attempt to identify who wrote a particular document by identifying this style, known as the author's \textit{idiolect} \citep{george1990whose}. Such models have proven useful for identifying who wrote historical documents \cite{holmes1995federalist}, identifying sockpuppet accounts \cite{solorio2013case,weerasinghe2022using}, and identifying forged documents \cite{silva2024forged}. Commonly, computational approaches to authorship attribution\footnote{The task of authorship attribution identifies which of $k$ authors wrote the document and can be framed as either a classification or, more recently, a retrieval (ranking) based problem \citep{stamatatos2009survey}.  } focus on the task of document representation, where models are trained to recover an author's stylistic signature from the text. As a result, modern approaches frame the task as whether two documents are written by the same author \cite{tyo2022state}, learning to represent documents by the same author so that they are nearby in the document space \cite{boenninghoff2019similarity,luar_2021}. Such models implicitly assume that authors are best modeled as documents, with the attribution task reduced to whether a new document's representation is near  the author's other documents. However, such approaches ignore systematic variation  and that an author's own idiolect might naturally have multiple substyles, depending on  context. Here, we approach attribution by learning a model of the author themselves---rather than just documents---to capture an author's natural stylistic diversity.

Authorship attribution and verification models have traditionally focused on identifying distinguishing and systematic stylistic differences in how an author writes documents \cite{stamatatos2009survey}. In forensic linguistics, stylometry, and stylometrics, style is represented using function word frequencies or manually curated syntactic patterns \cite{garcia2006function,brocardo2013authorship,tschuggnall2014enhancing}. More recent work has focused on using contrastive learning methods to learn document representations where documents by the same author are located near each other \cite{boenninghoff2019similarity,fincke2024separating}. In general, approaches project a document into a latent style space and identify an author using the authorship of documents nearby in that space. However, as seen in Figure~\ref{fig:intro-fig}, such approaches can overlook the regularity and distribution of an author's style, where a single author may have multiple distinct styles that can cause a document-level approach to misattribute authorship.

Here, we aim to learn representations of authors that explicitly model any natural variation. Our study is motivated by this prior work in showing that authors are sensitive to context in how they communicate \cite{kestemont2018overview,halvani2016authorship,sundararajan2018represents}, which strongly suggests that authors may have regular but dissimilar styles according to context. To study idiolectic variation, we learn models for 40K authors using 3M documents across four diverse corpora. Using four different models for authorship representation, we demonstrate that modeling the author leads to significant performance gains. Using our most accurate model of authorship, we then quantify how authors vary in their stylistic behavior in different contexts---i.e., when do authors write most like themselves? In four tests across audience, time, context, and genre, we demonstrate that authors reliably shift styles.

Our work has the following contributions.
First, we demonstrate that modeling the author's systematic style variation yields a significant performance improvement, regardless of how a document's style is represented.
Second, downstream analyses demonstrate that context is critical for accurate attribution: (1) authors vary more in style when writing across diverse contexts, (2) in settings with strong genre norms, authors are most distinctive in contexts where they have more stylistic flexibility, (3) in most settings, an author's style changes as they age, drifting from their earlier forms, and (4) surprisingly, in a negative result, authors do not shift their style when writing replies to another author. These results emphasize that the "where" of a text is essential to accurate attribution. All code will be released (CC-by-SA 4.0).

\section{Recognizing an Author's Style}
\label{sec:literature}

This work touches on multiple branches of research in NLP and Linguistics, focusing on quantifying an author's writing style.%

\noindent \textbf{Idiolect}
Authorship attribution depends, in part, on an author having a distinct writing style, or idiolect. In forensic linguistics, the idiolect is viewed as multifaceted--not just word usage, but also includes features such as syntactic and discursive patterns \citep{turell2011use}, and is thought to be persistent over time. However, two polarized views of idiolect have been formalized in the linguistics literature \cite{barlow2013individual,sep-idiolects}. \citet{labov2011exact} argued that idiolects can be understood as a reflection of the grammar of the speech community. Under this view, all idiolects can be reduced to the different combinations of sociolects (regular  variation  associated with a community). 

A contrasting view holds that individual languages should be primary, such that sociolect emerges from the grouping of idiolects \cite{hudson1996sociolinguistics}. Yet, open questions remain as to how to characterize idiolect systematically--- beyond dismissing it as idiosyncratic patterns---and how to separate idiolect from sociolect, if that distinction exists. While prior studies suggest that idiolect is stable over a short time period \cite{coulthard2004author,barlow2013individual,mojedano2024epistemic}, little is known about the temporal variation of idiolect over a longer period. Given the known influence of social conventions of genre and register on language use \citep[e.g.,][]{bischoff2020importance}, to what degree do these conventions constrain the variation within idiolects \cite{sep-idiolects}? Understanding idiolect variations has practical implications for stylometry and forensic linguistics as well as for the ontology of language itself \cite{carden1970note}.

\noindent \textbf{Author Attribution Models}
Computational models of authorship attribution depend on some explicit or implicit representations of the style of a document \citep{neal2017surveying}. Early computational work was focused on explicit linguistic features that are highly interpretable, including function words \citep{mosteller1963inference,garcia2006function}, character n-grams \cite{sapkota2015not}, and syntactic structures \cite{tschuggnall2014enhancing}. \citet{gill2011stylometric} proposed learning a Dirichlet prior over distributions of only function words to compare authors; while showing the benefits of adding structure, this approach was only tested with two authors and 14 documents. Despite these methods' early successes, they have been superseded by more powerful deep learning models \cite{Pardo2019OverviewOT,stamatatos2022overview,bevendorff2024overview}. %

There are two major directions in the deep learning approach with implicit authorship representations. Prompting-based work directly uses LLMs to identify authorship through text-only instructions and examples, with promising results \cite[e.g.,][]{huang-etal-2024-large,hu-etal-2024-bayesian}. However, this approach is limited to a small number of authors due to LLMs' context window sizes and cost. Retrieval-based approaches typically fine-tune an LLM to generate dense vector representations of style through contrastive learning, enabling efficient ranking over tens of thousands of authors \cite[e.g.,][]{boenninghoff2019similarity,luar_2021,zhuIdiosyncraticNotArbitrary2021,wang2023can,fincke2024separating}. 

Currently, most authorship attribution modeling techniques model the documents and do not explicitly parametrize a representation of the author. 

One example of a simple author modeling approach comes from \citet[]{zhuIdiosyncraticNotArbitrary2021}, which leverages document representations to build a Gaussian distribution at the author level. However, there is no notion of whether the distribution is a strong fit for the model. Further, this style of modeling is rigid, causing problems for authors who have a more complex idiolect. 

We aim to bridge this gap by building author models based on document representations of the text. We explore many different representations, as the strength of the representation is crucial in building performant models. Additionally, we analyze various modeling techniques for authors and enable modeling flexibility by leveraging a dev set for each author. This approach allows us to build models tailored to a given author's idiolect. 

\section{Datasets}
\label{sec:datasets}

To study author idiolect across diverse contexts and styles, we curate English-language data from four sources: comments from Reddit \cite{baumgartner2020pushshift}, literary books, emails from the public GMANE archive \cite{stein:2020o}, and Wikipedia \citep{israeli2025million}. These sources contain sufficient numbers of documents per author to model intra-author variation in different contexts and across time. We describe the preprocessing and data preparation steps next.

\noindent \textbf{Preprocessing}
For all data sources, we impose two general requirements on authors regarding their document lengths and the total number of documents. First, documents need to be long enough to evoke a specific style, and therefore we require at least 100 words; this limit is on the lower end for length in past work \cite{boenninghoff2019similarity} and ensures that the task is more realistic and challenging. Second, to enable a sufficient number of documents for downstream analyses, we retain only authors who have 100 or more documents, up to 250 total for the author. We collect 10K authors meeting both requirements from each source. See Appendix~\ref{app:sampling} for full details. 

\noindent \textbf{Reddit}
A large amount of content on Reddit comes from bots. To avoid modeling non-human users, we removed these accounts using a large curated stoplist and filtering rules (e.g., usernames ending in ``Bot").  Using the remaining accounts, we constructed a balanced subset of 10K authors by evenly sampling 100 authors from each combination of (1) total years of Reddit participation (binned into integer buckets from 0 to 9 years), and (2) the number of unique subreddits they had posted in (binned into ranges: 1–5, 6–10, ..., 46–50). This stratified sampling ensures  a diverse group of authors in both longevity and community participation.

\noindent \textbf{Books}
Books represent a unique genre due to their length and register constraints, which influence the style and narrative. To model this influence, we sample 10K novels converted to plain text \citep{gao2020pile},  partition their content at the paragraph level, and retain only paragraphs with 100+ words.  

\noindent \textbf{GMANE}
The public GMANE archive contains emails and replies on mailing lists for a variety of topics \citep{stein:2020o}. We perform document sampling at the author level with respect to emails and replies with the technique described in Appendix~\ref{app:sampling}. This email-and-reply setting allows us to later measure the effect of style influence.

\noindent \textbf{Wikipedia}
Wikipedia contains diverse contributions from authors on its article and talk pages. We select Wikipedia authors from the Million Author Corpus \citep{israeli2025million} and again perform the metadata sampling described in Appendix~\ref{app:sampling}.

\section{Modeling Author Style Variation}
\label{sec:models}

We evaluate four types of parametric representations of authors derived from the style of their documents. To control for a potential interaction between style representation and author model, we consider six different style representations, two interpretable and four neural. We use document representations to decouple the problem of learning stylistic features from modeling author-specific patterns; by first encoding style, author models operate on stable, reusable, topic-deconfounded representations rather than raw text. 

\subsection{Author Models}

\noindent \textbf{Nearest Neighbor}
As a representation of current practice, we evaluate a nearest-neighbor model to calculate $p(a_i|d_j)$ for author $a_i$ and document style representation $d_j$.\footnote{We elide subscripts in this notation as $p(a|d)$ when it is clear what the author and document refer to.} Given some new document $d_j$, we compute $\arg\max_ {d_k \in \text{docs}(a_i)} \text{cos}(d_j, d_k)$ and then $p(a_i | d_j) = \frac{cos(d_j, d_k^{a_i})}{\sum_{a_j \in \text{Authors}} cos(d_j, d_k^{a_j}) }$. In essence, this model-free approach relies on the similarity of the nearest neighbor to identify authorship. 

\noindent \textbf{Centroid}
\citet{zhuIdiosyncraticNotArbitrary2021} proposed an alternative and simple model where each author is represented as a single Gaussian distribution parameterized by $\mathcal{N}(\mu_a, \Sigma_a)$, where $\mu_a$ and $\Sigma_a$ denote the mean and covariance matrix specific to author $a$. This model captures the overall distribution of an author's style. We compute $p(a|d)$ for each author and then normalize the scores with a softmax.

\noindent \textbf{Gaussian Mixture Models}
Authors are thought to have distinct styles depending on context, and therefore we propose using a Gaussian Mixture Model (GMM) parameterized by $\sum_{k=1}^{K} \pi_{a,k} \, \mathcal{N}(\mu_{a,k}, \Sigma_{a,k})$, where $\pi_{a,k}$ are the mixture weights, and $\mu_{a,k}$ and $\Sigma_{a,k}$ are the mean and covariance of the $k$-th component for author $a$. Such a model can naturally represent an author's multiple distinct and consistent styles. However, a GMM requires us to define the number of components, $k$, for the GMM. 

Rather than force all authors to have a fixed $k$, we use validation data to optimize $k$ per author. Like the centroid, we normalize $p(a|d)$ across all authors.

For performance, we constrain the GMM covariance matrices to be diagonal. The style representations (described below) are inherently correlated, so we orthogonalize via principal component analysis (PCA). This orthogonalization allows us to avoid computing the full covariance matrix and reduces the complexity of fitting models for all authors from $O\!\left(T(nkd^2 + kd^3)\right)$ to $O(nd^2 + d^3) + O\!\left(Tnkd\right)$, where $n$ is the number of documents, $d$ the embedding dimension, and $T$ the number of EM iterations. 

Appendix~\ref{app:orthogonalization} details PCA's time complexity and performance gains.

\noindent \textbf{RoBERTa}
Pretrained language models (PLMs) can capture stylistic information \cite{hayati-etal-2021-bert}. Prior work on authorship verification used BERT-based classifiers to predict which of $n$ authors wrote the text \citep[e.g.,][]{fabien2020bertaa}, though such models are often trained for small values of $n$ (e.g., 100). Here, we follow this approach and fine-tune a RoBERTa model to predict authorship directly from the text, rather than using intermediate style representations of the documents, which creates an implicit model of authors' styles. %

\subsection{Document Stylistic Representations}

The author models are fit as distributions of their documents' style representations. Several methods have been proposed for representing document style in authorship attribution, and we employ six distinct representations. (1) Style is represented as the frequency of the 67 lexical and syntactic features from Biber's Multidimensional Analysis framework \citep{biber_1988} and 29 additional features from \citet{clarke_dimensions_2017} and \citet{clarke_functional_2019}. These interpretable features characterize style through rule-based linguistic markers. For scalability, we use NeuroBiber \citep{alkiek2025neurobiberfastinterpretablestylistic} to estimate these features. 
(2) An alternative to the low-level style representation of Biber is LISA \citep{lisa_2023}, which uses a PLM to generate interpretable embeddings aligned to stylistic dimensions like tone, emotional expression, and linguistic patterns.  These higher-level features capture a complementary view of style, though one not directly optimized for the task of authorship attribution. 
(3) The LUAR \citep{luar_2021} model is a PLM trained via contrastive learning to represent documents by the same author with similar embeddings. The model is one of the first accurate neural style embedding models.
(4) The PCM model \cite{kim2025leveraging} is the current state of the art for multilingual neural style embedding. 
(5) The StyleDistance model \cite{patel2025styledistance} was trained using synthetic data to maximally de-confound topic and style.
(6) The SADIRI model  \cite{fincke2024separating} uses adaptive exemplar filtering for contrastive learning to create representations that perform well in cross-genre settings.

\section{Experiment: Author Attribution}
\label{sec:experiment}

Here we assess whether explicitly modeling an author's stylistic variation improves performance.

\noindent \textbf{Setup}
We fit each type of author model on the full dataset  (40K authors), reporting performance by dataset. For the GMM, we optimize $k$ using a grid search to fit models in $k \in [1,7]$, selecting the $k$ that maximizes the probability on the author's development data.
Details of RoBERTa training, GPU compute, and author-model runtime are in Appendix~\ref{app:training_details}. 

Models are evaluated for authorship attribution using both Success@8, which is a more recent common metric that captures whether the true author is among the 8 most likely candidates \citep[e.g.,][]{agarwal2025cross,israeli2025million}, and $p(a|d)$. Both metrics offer complementary insight: Success@8 reflects the practical performance of an authorship attribution system, while $p(a|d)$ allows us to assess where a model can identify fine-grained changes to idiolect that may not ultimately affect the  ranking. Error bars throughout this paper show 95\% bootstrap confidence intervals from 1,000 resamples.

\begin{figure*}[t]
  \centering
  \includegraphics[width=0.48\textwidth]{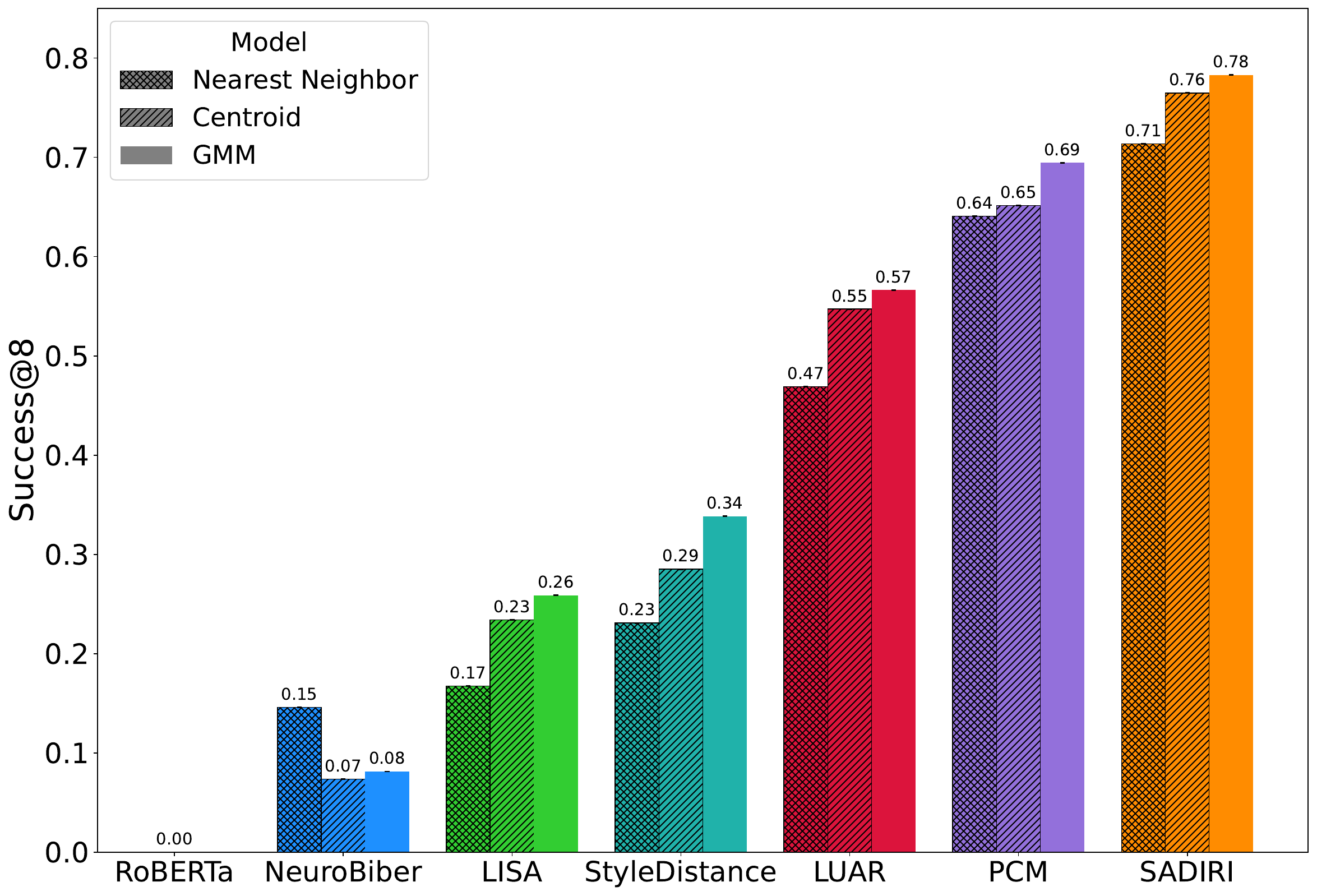}
  ~
  \includegraphics[width=0.48\textwidth]{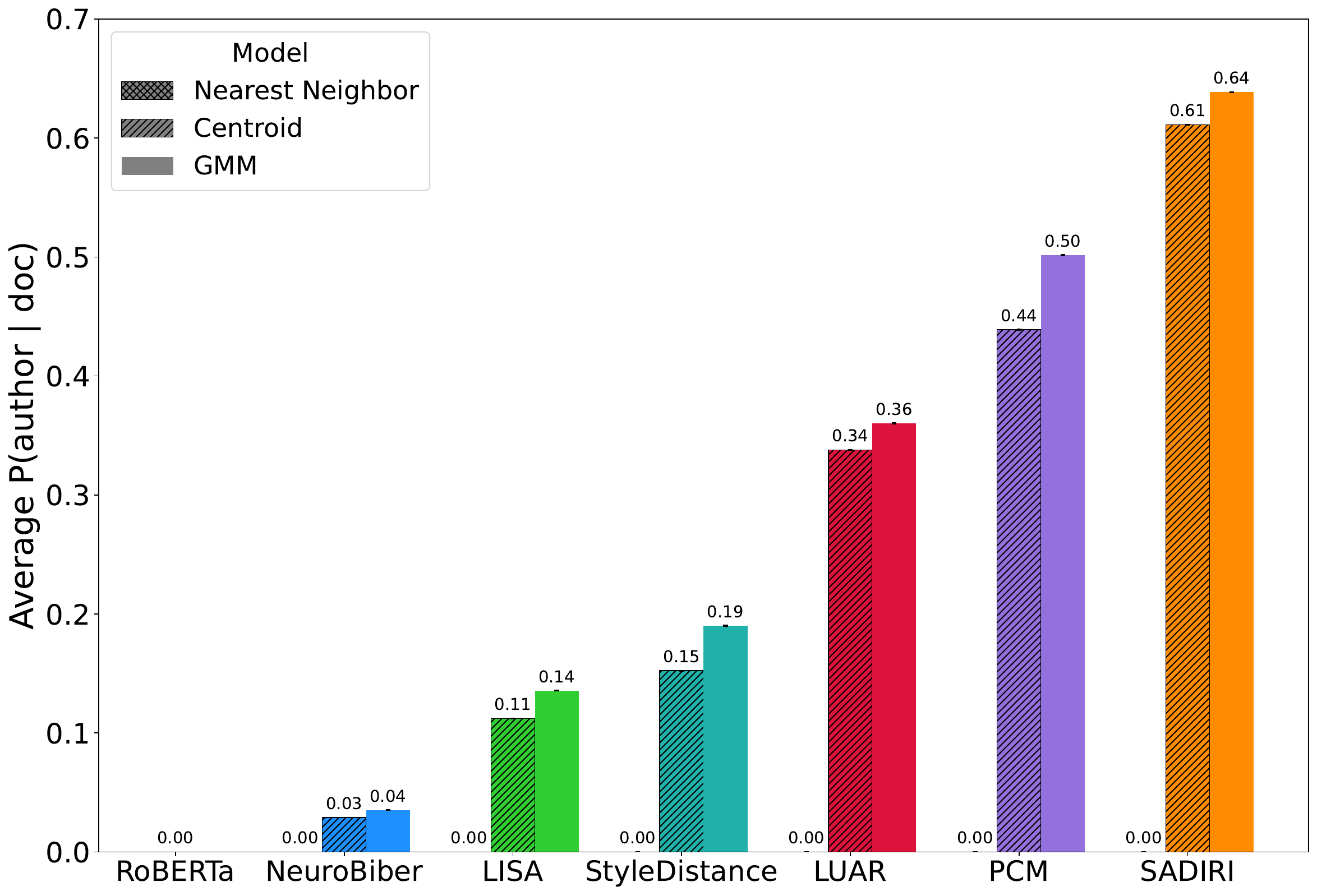}
  \caption{ Performance at attribution documents of 40K authors for different representations of authors and styles, shown as Success@8 on the left and $p(a|d)$ on the right.  Representing authors as mixtures of styles yields consistent performance gains for both metrics. Note that the probabilities/Success@8 for the Nearest Neighbor and RoBERTa model are non-zero but round to 0.00. }
  \label{fig:experiment_comparison}  
\end{figure*}

\noindent \textbf{Results}
Regardless of how we represented style, modeling authors as mixtures of consistent styles with a GMM yields substantial and significant performance gains for both metrics, as seen in Figure~\ref{fig:experiment_comparison}; see Appendix~\ref{app:performance_metrics} for the corresponding numerical results. Appendix~\ref{app:source_analysis} further breaks down performance by source. Further, the performance gains are largest for the highest-performing style representations, showing that better style representations provide additional fidelity to model an author's distinct mixture of styles. These results also highlight the importance of style modeling: while using a similar number of parameters as the LUAR and PCM models, the implicit representations of authorship in the RoBERTa model (Figure~\ref{fig:experiment_comparison} left) are likely insufficient for accurate authorship attribution without additional training approaches like the contrastive learning used by the four neural embeddings. Surprisingly, the commonly used approach of using nearest neighbors also yields very low performance (with the exception of NeuroBiber), suggesting that even adding a simple author model like a Centroid can produce immediate gains. 

\noindent \textbf{Is the GMM just learning topical variation?}
Two of the state-of-the-art models we use, SADIRI and StyleDistance, are designed to be robust to topic through maximizing cross-genre performance or training on synthetic data, respectively. The GMM improves performance significantly with both representations, suggesting the potential topic confound is not responsible. Appendix~\ref{app:heldout_subreddit} further tests this with held-out Reddit communities.

\noindent \textbf{How many components do authors need?}
Our results show that modeling authors as one or more Gaussians yields consistent gains. Examining the distribution of optimal GMM components ($k$), we find that  21\% of author models have a single component---equivalent to the centroid model---and 61\% of author models have between 2-4 components. Surprisingly, 18\% of authors benefit from a 5+ component GMM. The full distribution can be found in Appendix~\ref{app:components} as well as the breakdown per style representation type. This result deviates sharply from current practice in authorship attribution but supports prior work on style that suggests authors do have multiple distinct styles \cite{kredens2021100,grant2018resources}. 

\noindent \textbf{How many documents are needed to see intra-author variation?}
Our prior setup used 50 documents per author to fit a GMM and an additional 25 documents to select the optimal $k$. However, in some settings, fewer documents may be available, raising the possibility that a GMM would be unneeded---i.e.,  fewer documents could provide insufficient evidence of systematic variation in an author's style. To test this, we follow a similar setup where we train the same GMM models on varying numbers of training documents, using the best-performing SADIRI style embedding. To test for saturation, we include an additional 30 documents per author, still choosing  $k$ according to the highest log-likelihood on a held-out validation set. 

The results show that even with a few training documents (e.g., 10), there is benefit to GMM-based modeling as the average optimal $k$ is above 2. However, once we get to around 20 documents, we do not see an increase in the number of components for authors, indicating that additional documents do not increase complexity.

\section{Does contextual diversity influence stylistic consistency?}
\label{sec:contextual_diversity}

Authors may naturally vary their style when communicating with different audiences in order to mirror community norms.  While certain stylistic features may remain relatively stable across topics and communities, authorship attribution becomes substantially more challenging under cross‑domain or cross‑genre variation. Indeed, previous work  highlights how attribution systems struggle when texts differ in topic, even when authored by the same individual \citep{kestemont:2019}. However, there is evidence that stylometric markers retain their informativeness even across multiple topics \citep{litvinova2018assessing}. Here, we measure this phenomenon on Reddit by using subreddits as a proxy for community, topic, and genre, asking to what degree authors are predictable relative to their contextual diversity. 

\noindent \textbf{Setup}
We reuse the SADIRI embedding GMM models for the 10K Reddit authors from Section~\ref{sec:experiment}. We then estimate the $p(a|d)$ using the authors' 1.1M test documents and stratify according to the number of unique subreddits the author writes in.

\begin{figure}[tbp]
  \centering
  \includegraphics[width=0.47\textwidth]{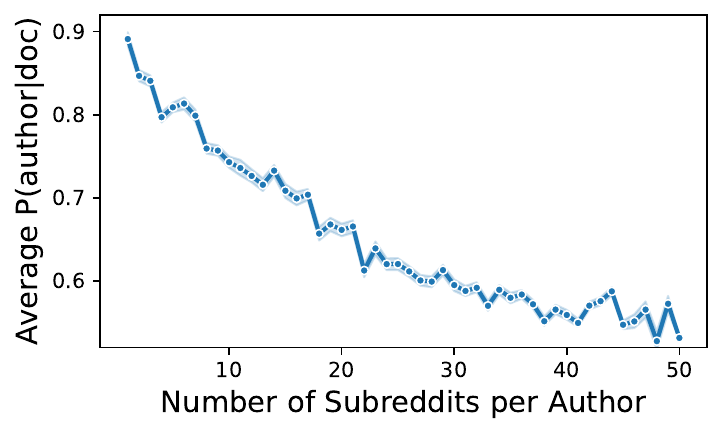}
  \caption{As the number of communities that an author writes in increases, we see a decrease in stylistic consistency. Authors that write in a large number of subreddits write in their own style significantly less than authors who are focused in a few.}
  \label{fig:subreddits}
\end{figure}

\noindent \textbf{Results}
Authors with participation in more communities have less distinctive style (Figure~\ref{fig:subreddits}). Indeed, authors who post in only a few subreddits maintain a highly distinctive writing style, with $P(a|d)$ values near 0.85. However, as the number of subreddits increases, this distinctiveness decreases rapidly. Such a trend has two likely causes. First, authors likely adapt to the increasing number of communities, each with different practices, creating a more varied stylistic profile that is more difficult to model with limited data. Second, while our style representations are trained to be robust to content or topic, many authorship attribution models incorrectly learn author-topic associations \cite{wegmannSameAuthorJust2022}, which can lead to worse performance when authors write in multiple topics.

\section{Do narrative conventions constrain author style?}
\label{sec:narrative_conventions}

Narrative conventions significantly shape and shift authorial style. \citet{montfort-etal-2021-using} show that controlling referring expressions, such as how characters or locations are introduced, yields distinct narrative styles. Similarly, \citet{schwartz-etal-2017-effect} find that writing task framing—like continuing a story versus ending it—results in measurable stylistic variation. These works suggest narrative constraints influence expression beyond an author's baseline tendencies.
Where then in novels---if anywhere---are authors least constrained by these conventions?  We investigate these findings by evaluating author style at different segments of long-form documents using the books data. 

\noindent \textbf{Setup}
From the data preprocessing step, books are segmented into chunks that are tagged based on their relative position within the book. We group these documents into deciles according to their relative position within the book. Then, we measure $p(a|d)$, aggregated on each of these deciles for all 10K authors and test documents. 

\begin{figure}[tbp]
  \centering
  \includegraphics[width=0.47\textwidth]{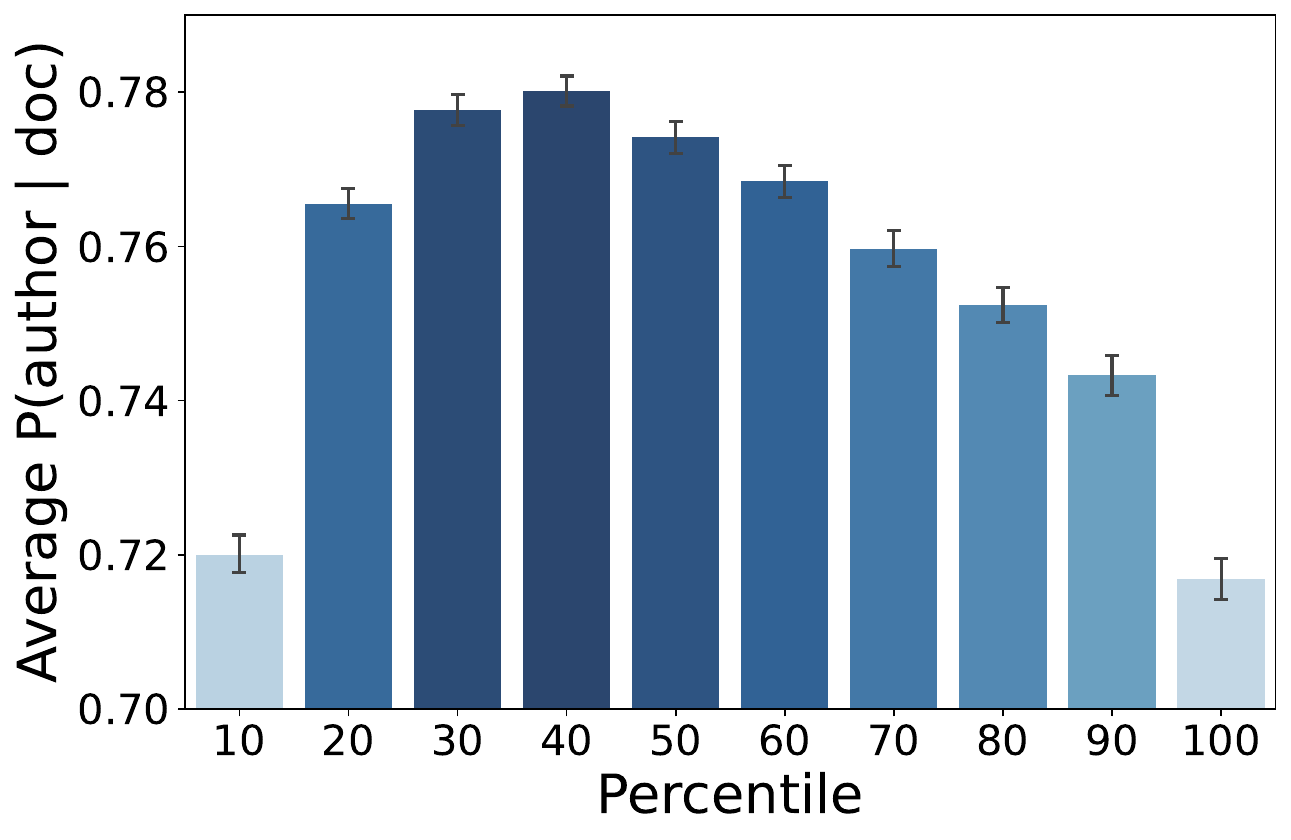}
  \caption{In longer-form documents, authors tend to write least like themselves in the beginning and ending portions. Authors write most like their style in the middle of the book, specifically just before the middle.}
  \label{fig:books_consistency}
\end{figure}

\noindent \textbf{Results}
Figure~\ref{fig:books_consistency} shows a clear pattern: authors are most stylistically consistent in the middle of their books, and least consistent at the beginning and end. This suggests that the conventions of writing introductions and conclusions suppress an author's own style, likely due to functional constraints; authors often open with more stylistically conventional exposition to orient the reader and close with summary or resolution. The middle sections, which tend to carry the core narrative, allow for more sustained and unconstrained expression of authorial voice. Our results suggest that when attributing authorship in long-form text, practitioners must account for genre and discourse constraints, as not all passages are equally informative about authorial style \citep{mosteller1963inference}.

\section{How consistent are authors over time?}
\label{sec:temporal}

An individual's language changes as they age, which raises the question of to what degree their style also changes. Previous work on shorter time frames has shown evidence that author idiolects tend to stay relatively stable \citep{blas2023individual,barlow2013individual}. Yet, the few sociolinguistic studies on longer time frames with only a few individuals found that their sociolect does change \citep{rickford2013girlz,sankoff2019age}, suggesting that idiolect might as well. 

\noindent \textbf{Setup}
New GMM models are trained and evaluated on the SADIRI style embeddings using a chronological data split. The first 50 documents per author (temporally) in Reddit, GMANE, and Wikipedia are used to fit the GMM. The $k$ value is chosen by validating on the next 10 documents. We evaluate on the remaining documents. For each test document, we measure $p(a|d)$ relative to the time elapsed since that author's last training document. Results are pooled at the month level for the  37,714 Reddit, 63,372 GMANE,  156,918 Wikipedia article, and 26,750 Wikipedia talk page documents. All authors are equally present in all windows.

\begin{figure}[tbp]
  \centering
  \includegraphics[width=0.43\textwidth]{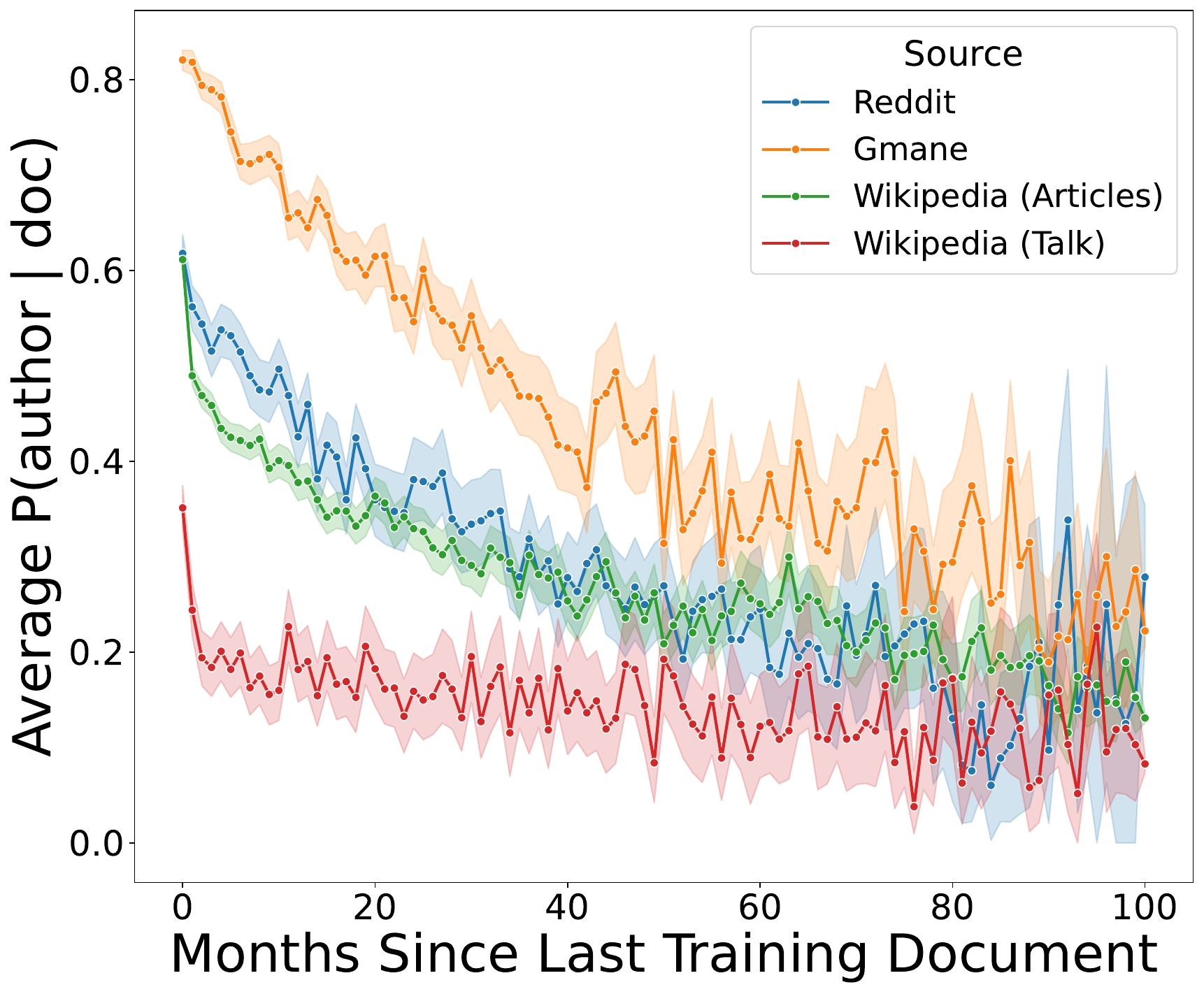}
  \caption{As time progresses, authors deviate more and more from their initial style. This indicates that author style changes significantly over time, and author models trained in the past may not be as effective as authors continue to write. This effect is particularly noticeable on Reddit and GMANE, which both show a steep decrease, whereas Wikipedia drops initially but then continues with only a small decline.}
  \label{fig:temporal}
\end{figure}

\noindent \textbf{Results}
Figure~\ref{fig:temporal} shows that author predictability declines as the time gap between training and test documents increases, with Pearson correlation tests significant across all four genres ($p<0.01$). $P(a|d)$ drops sharply within 24 months, especially on Reddit and GMANE. This decrease suggests that authors' underlying idiolectic styles change over time, becoming dissimilar from their earlier style. 
Wikipedia has a more nuanced decrease in predictability. Both Wikipedia sources see an overall decrease, but the article pages follow the same trend as Reddit and GMANE, while the harder-to-attribute talk pages see only a small decrease. 

\noindent \textbf{Why are authors less predictable over time?}
The drop in predictability could be due to multiple changes in stylistic behavior. Here, we test three hypotheses: \textbf{H1} The author becomes less stylistically consistent over time and is unable to be accurately identified, \textbf{H2} The author adopts a new, \textit{additional} substyle that is not present initially, \textbf{H3} The author changes to a new, distinct style with equal complexity. 

To test these hypotheses, we employ a sliding window approach that segments each author's chronological document series into discrete temporal windows, allowing us to compare their styles at different points in time. Briefly, we use 50 consecutive documents to fit a GMM and compare the documents in other windows using the log-likelihood; full details are in Appendix~\ref{app:temporal-analysis}.

\begin{figure}[tbp]
  \centering
  \includegraphics[width=0.47\textwidth]{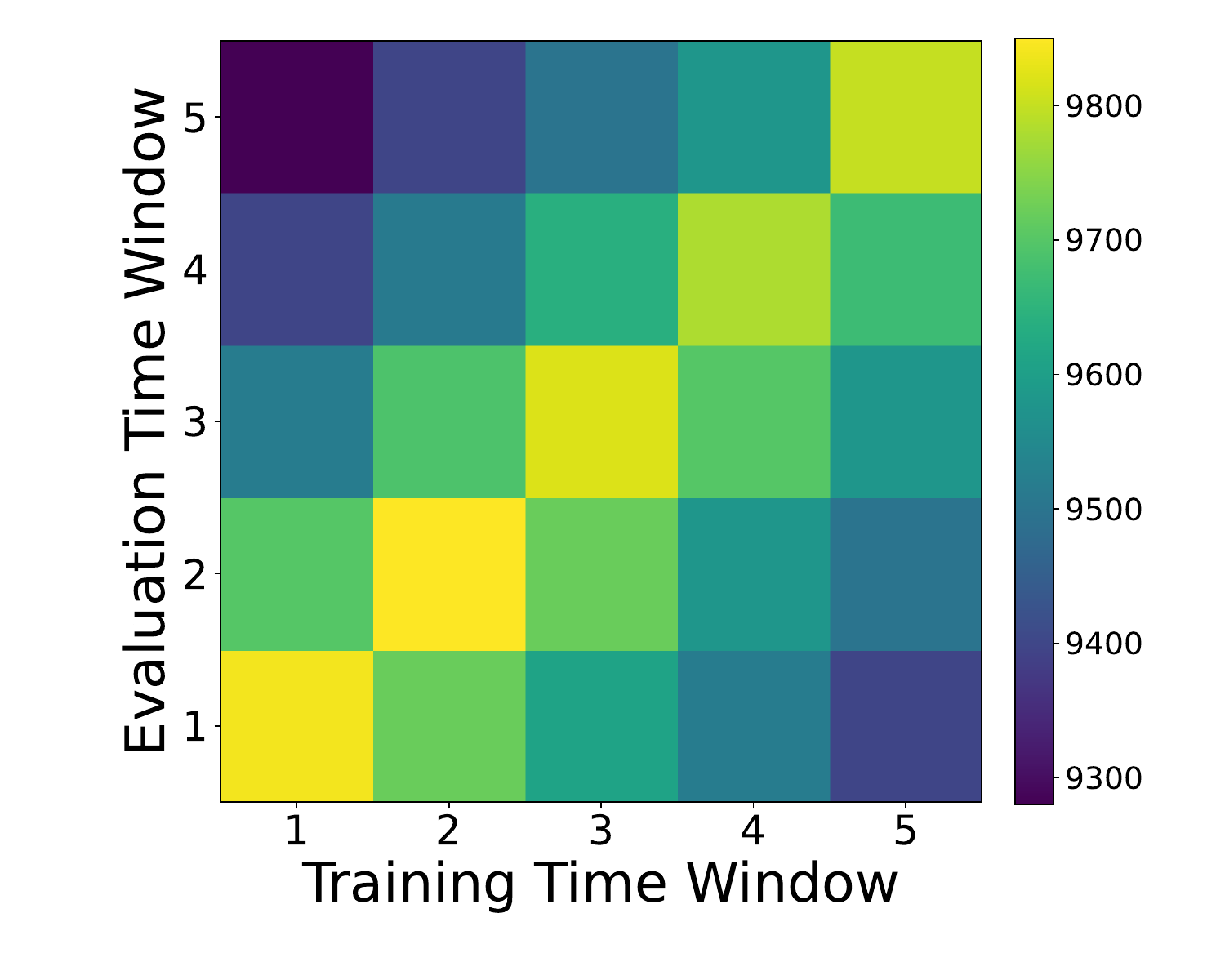}
  \caption{Heatmap of average log-likelihood for pairs of model window and data window. Performance is highest along the diagonal and declines with increasing separation, indicating distributional drift over time. 
  }
  \label{fig:time_heatmap}
\end{figure}

Our results support one of the three hypotheses. First, we find evidence for consistent evolution of a stable but distinct idiolect; Figure~\ref{fig:time_heatmap} shows that held-out documents within a window are reliably identified as an author's (the diagonal) but this identification drops over time. Thus, we do not find support for H1, which would have a decreasing likelihood over the diagonal. Second, we plot the mean number of components (Appendix Figure~\ref{fig:temporal-components}) showing that authors do not add or lose style components, maintaining complexity. Finally, Figure~\ref{fig:temporal-lag} shows that temporal degradation is driven more by location than variance drifts. This indicates stylistic displacement rather than inconsistency. Together, these results support H3: author style shifts over time without changing in complexity.

\section{Do authors shift style in conversation?}
\label{sec:conversation}

Linguistic accommodation theory suggests that interlocutors often adjust parts of their language to match that of their conversational partners \citep{giles2008communication} , and methods for studying accommodation have used function words \citep{litterer2025coordinating}, suggesting this effect could be from changes in idiolect. Therefore, we ask to what degree authors retain their own style when replying versus adopting the idiolect of the other?

\noindent \textbf{Setup}
Here, we re-use the trained models on the SADIRI embeddings, evaluate on GMANE data in the test set, and measure how $p(a|d)$ changes depending on which type of message an author is sending. To control for differences between authors, we restrict our analysis to pairs of authors $a_1$, $a_2$ where each author has made an initial email (a post) and the other author has been the first reply to that post. Using the 1.3M emails in our test data, we identify 956 post-reply pairings and 1109 unique authors that engage in these conversations. 

Let author $a_1$ write a reply $r$ to a post $p$ by author $a_2$, and let $p_1$ denote a distinct baseline post written by $a_1$. Using posterior authorship probabilities $P(a \mid d)$, we define two deltas:
$
\Delta_{\text{self}} = P(a_1 \mid r) - P(a_1 \mid p_1)
$
and
$\Delta_{\text{parent}} = P(a_2 \mid r) - P(a_2 \mid p_1).
$
$\Delta_{\text{self}}$ measures whether the reply remains identifiable as $a_1$’s style, while $\Delta_{\text{parent}}$ measures movement toward the parent author $a_2$. Negative $\Delta_{\text{self}}$ indicates self drift and positive $\Delta_{\text{parent}}$ indicates accommodation. 

\noindent \textbf{Results}
When authors reply, they become slightly more identifiable, but show no meaningful accommodation towards the parent author. This suggests that replies may elicit richer stylistic features for the author without being influenced by the parent author's style. Appendix Figure~\ref{fig:gmane} shows that authors maintain their idiolect rather than converging toward the style of the author they are replying to. Thus, replies remain a rich signal of idiolect. 

\section{Conclusion}
\label{sec:conclusion}
In this study, we seek to model the style representations of \textit{authors}, rather than those of \textit{documents}, showing that such an approach leads to better representations of authorship. With large-scale analyses of 40K authors and millions of documents across four genres, we demonstrate the importance of the writing context. Authors exhibit stylistic variations across conversation contexts, genre norms, and time. Our findings provide a more nuanced understanding of idiolects, highlighting the need to account for context-dependent variation while modeling the invariant aspects of style. Beyond improving authorship verification, our findings can inform sociolinguistics and the development of fairer, more context-aware NLP systems.  

\newpage

\section*{Limitations}
\label{sec:limitaitons}

First, we acknowledge that it is possible that our model is learning a shortcut for idiolect, such as topic. Although each of the embeddings we use attempts to disentangle topic, there is no perfect way to do this. StyleDistance provides a strong signal regarding our modeling setup because it is explicitly trained on synthetic data to de-confound topic and style. Figure~\ref{fig:experiment_comparison} shows that StyleDistance has one of the strongest performance gains from our setup, which makes us more confident that we are modeling stylistic features rather than topical ones. However, author models may still partially encode topic or community-level regularities rather than purely idiolectal variation. Although the held-out subreddit evaluation in Appendix~\ref{app:heldout_subreddit} suggests the model retains substantial stylistic signal under cross-topic generalization, performance does decrease in this setting, indicating that some contextual confounding may remain even with an embedding optimized for cross-genre style representation.

Although we find a strong performance gain from our methodology, we acknowledge that there are limitations to our analysis. First, we leverage a moderate number of documents to build our author models (50), which may not always be feasible. We still believe our approach is beneficial even with a smaller number of documents, as shown in Appendix Figure~\ref{fig:avg_k}, but further work would be needed to quantify the trade-offs relative to the number and length of documents available. Second, the model performance still varies depending on the setting in which the document was written. In Sections 6 to 9, we show many different contexts where an author appears to deviate from their own idiolect, as represented by our models. Additionally, in Appendix~\ref{app:wiki_genre} we see that authors who write consistently in one setting are not guaranteed to write consistently in another. Thus, while our authorship models do improve performance, these models are still not fully robust to the reality of contextual effects on writing and would be limited in performance depending on where and when the data used to train and evaluate the models was collected. 

\section*{Ethics Statement}
\label{sec:ethics}
Our study primarily focuses on the theoretical aspects of individual styles. We only made use of public datasets, and all authors in the datasets were properly anonymized to avoid any potential disclosure of real identities.  We acknowledge that the authorship verification technology is dual-use. While such technology can be used for forensic investigation and historical authorship attribution, malicious use can also potentially compromise the anonymity of writers and contribute to mass surveillance. We urge the responsible deployment of such technology. Our study shows the fluidity of individual styles, challenging the assumption that individuals' writing styles are invariant. Therefore, authorship verification technology should not be deployed in high-stakes scenarios without significant human oversight.  

\bibliography{main}

\newpage
\appendix

\section{Data Sampling}
\label{app:sampling}

To construct the full dataset, we sample 10K authors from each data source to build a final set of 40K authors. Although we set a lower bound on authors at 100 documents, some authors have a large number of documents, some even in the tens of thousands. Thus, we set an upper bound of 250 documents per author. If an author has more than 250 documents, we sample documents to follow the author's underlying metadata distribution. As an example, we analyze how the sampling would work for Wikipedia documents. Formally, let \(N_s\) be the number of documents in namespace \(s\) for a given author and \(N = \sum_s N_s\). We set
\[
k_s = \left\lfloor \frac{N_s}{N} \times 250 \right\rfloor,
\qquad
R = 250 - \sum_{s} k_s.
\]
We then allocate one additional post to each of the \(R\) namespaces with the largest fractional remainders:
\[
\frac{N_s}{N} \times 250 - k_s,
\]
This ensures that exactly 250 documents are sampled and that they follow the underlying author's namespace distribution as closely as possible.

Once the final set of documents is constructed, we create three splits: train, validation, and test. The distribution is 50 for training, 25 for validation, and $N$-75 for Test.
\\
We sample these documents with the intent to ensure that the splits follow the underlying metadata distribution. On average, we observed that authors had 128 test docs. The breakdown for test documents per source can be seen in Table~\ref{tab:test_counts}.

\begin{table}[htbp]
  \centering
  \begin{tabular}{|c|c|}
    \hline
    \textbf{Source} & \textbf{Test Count} \\
    \hline
    Reddit & 1,154,014 \\
    GMANE & 1,206,047 \\
    Wikipedia & 1,344,228 \\
    books & 1,421,921 \\
    \hline
  \end{tabular}
  \caption{All sources have a training set of 500K and a development set size of 250K, while test set sizes vary as shown. All sources have over 100 test documents per author on average.}
  \label{tab:test_counts}
\end{table}

\section{Source Based Analysis}
\label{app:source_analysis}

In this section, we investigate the differences between the four data sources. This can help us uncover phenomena occurring because of our modeling. 

\subsection{Performance Differences}

\begin{figure}[htbp]
  \centering
  \includegraphics[width=0.47\textwidth]{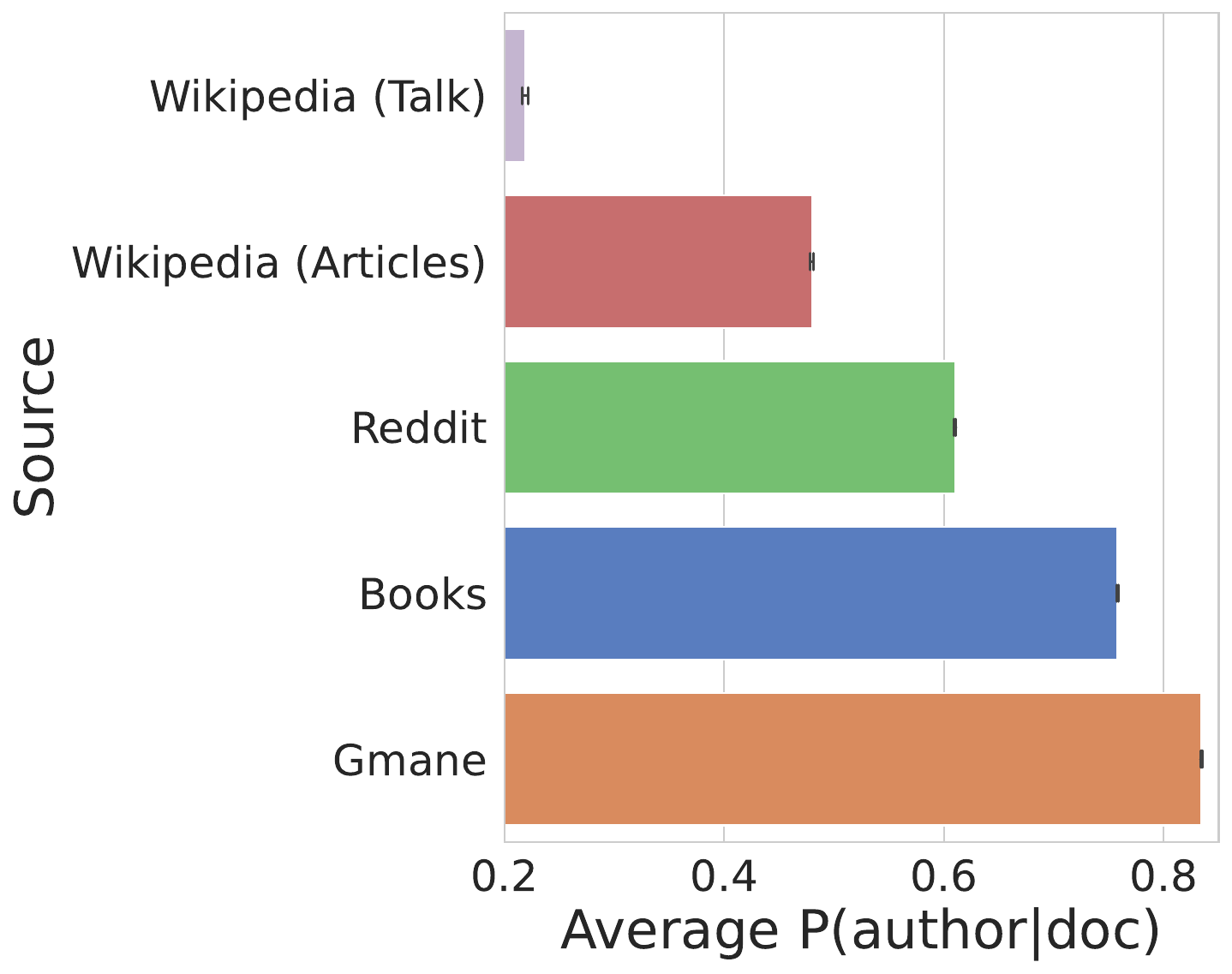}
  \caption{This figure shows consistency across the four data sources. Wikipedia has the lowest consistency by far, and the opposite can be said for GMANE. We hypothesize that this is a result of Wikipedia having more rules and structure, while emails are more informal and are used for day-to-day speech.}
  \label{fig:performance_by_source}
\end{figure}

To measure the performance by source, we simply took all of the testing documents and segmented them by their data source. The results, which can be seen in Figure~\ref{fig:performance_by_source}, uncover an interesting finding where authors from the GMANE data source are much more consistent than the other sources. Additionally, authors who write in Wikipedia talk pages have very low consistency. This may be attributed to the fact that emails are a richer stylistic source, as they are used for day-to-day messaging. On the other hand, Wikipedia talk pages usually are short edit proposals, making it difficult for authors to engage in their own idiolect.

\subsection{Change in Optimal Components}

\begin{figure}[htbp]
  \centering
  \includegraphics[width=0.47\textwidth]{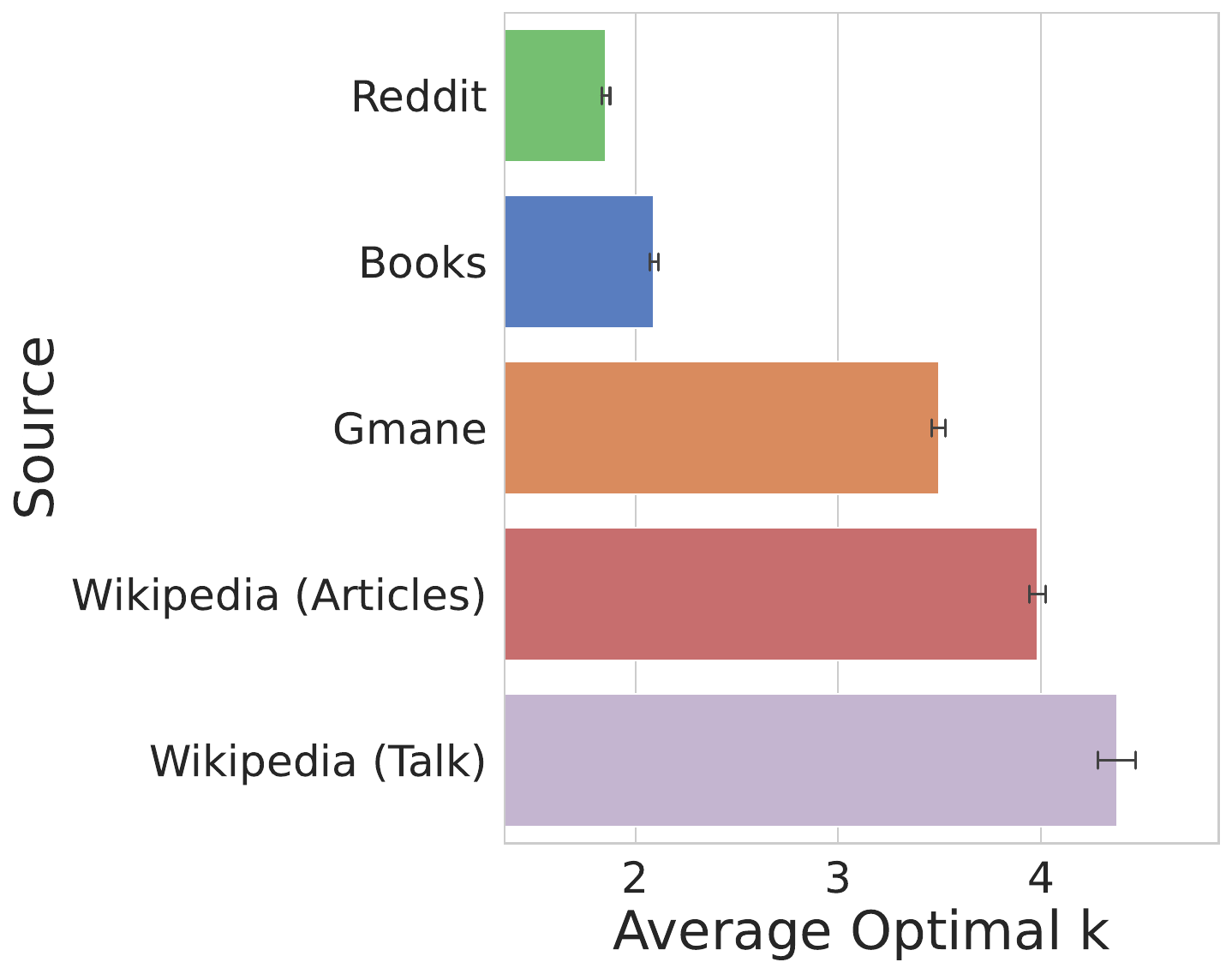}
  \caption{GMANE and Wikipedia have more complex idiolect distributions for their authors. Books and Reddit have comparable values, making it difficult to determine what causes authors to have more components. It seems plausible that GMANE has a higher number of styles because it can represent day-to-day and formal speech. However, Wikipedia usually follows a similar norm for its pages, so we would expect a lower number of styles.  }
  \label{fig:average_k_by_source}
\end{figure}

We also wanted to measure the change in components across the sources to see if some authors had more complex idiolect distributions than others. Figure~\ref{fig:average_k_by_source} shows that Reddit and books have a comparable number of styles while GMANE and Wikipedia have 0.5 and 1 more styles on average, respectively. This result paired with Figure~\ref{fig:performance_by_source} shows that more complexity does not mean higher consistency. Thus, there exists some tradeoff between complexity and performance, reinforcing the findings in~\ref{tab:k_distribution} that over half of authors have 2 or fewer components. From these findings, we can be more confident that the model is not learning a topic shortcut because Reddit would be much more topic diverse than books. 

\section{Held-Out Subreddit Evaluation}
\label{app:heldout_subreddit}

We evaluate cross-topic authorship attribution on Reddit by treating subreddits as proxies for topic/community. For each author, we construct author-specific splits over disjoint subreddit sampled documents. For our setup, authors must have at least 10 subreddits with at least 5 documents in each. Eligible subreddits are allocated 60/20/20 to train, dev, and test, with train and dev documents capped at 50 and 25 per author, respectively, while all documents from held-out test subreddits are retained. As a control, we also construct a random split with the same per-author train/dev/test document counts. Again, we use the most performant model: GMM with the SADIRI embedding. After applying data filters, we had 2\,669 authors retained, which are used for both setups. 

\begin{table}[htbp]
    \centering
    \begin{tabular}{lcc}
      \hline
      \textbf{Split} & \textbf{Success@8} & \textbf{$p(a\mid d)$} \\
      \hline
      Random & 0.751 & 0.573 \\
      Held-out & 0.684 & 0.493 \\
      \hline
    \end{tabular}
    \caption{Performance under random and held-out-subreddit splits for the cross-topic experiment.}
    \label{tab:cross-topic}
  \end{table}

The held-out subreddit evaluation shows that performance decreases when train/dev/test documents come from disjoint communities, but remains strong overall. Under the matched random-sampling control, the model achieves 0.751 Success@8 with mean $p(a|d)=0.573$. When train/dev and test documents are drawn from disjoint subreddit sets, performance drops to 0.684 Success@8 and mean $p(a|d) = 0.493$ over the same dataset.

The decline indicates that subreddit context contributes to attribution difficulty. Since subreddits differ in topic, audience, and norms, held-out-subreddit evaluation is a harder out-of-distribution setting than random document splitting. However, the drop is modest, suggesting that the model does not rely solely on subreddit-specific topical cues. The learned models retain substantial idiolectic signal in held-out communities.

\section{Number of Optimal Components}
\label{app:components}

\begin{figure}[tbp]
  \centering
  \includegraphics[width=0.47\textwidth]{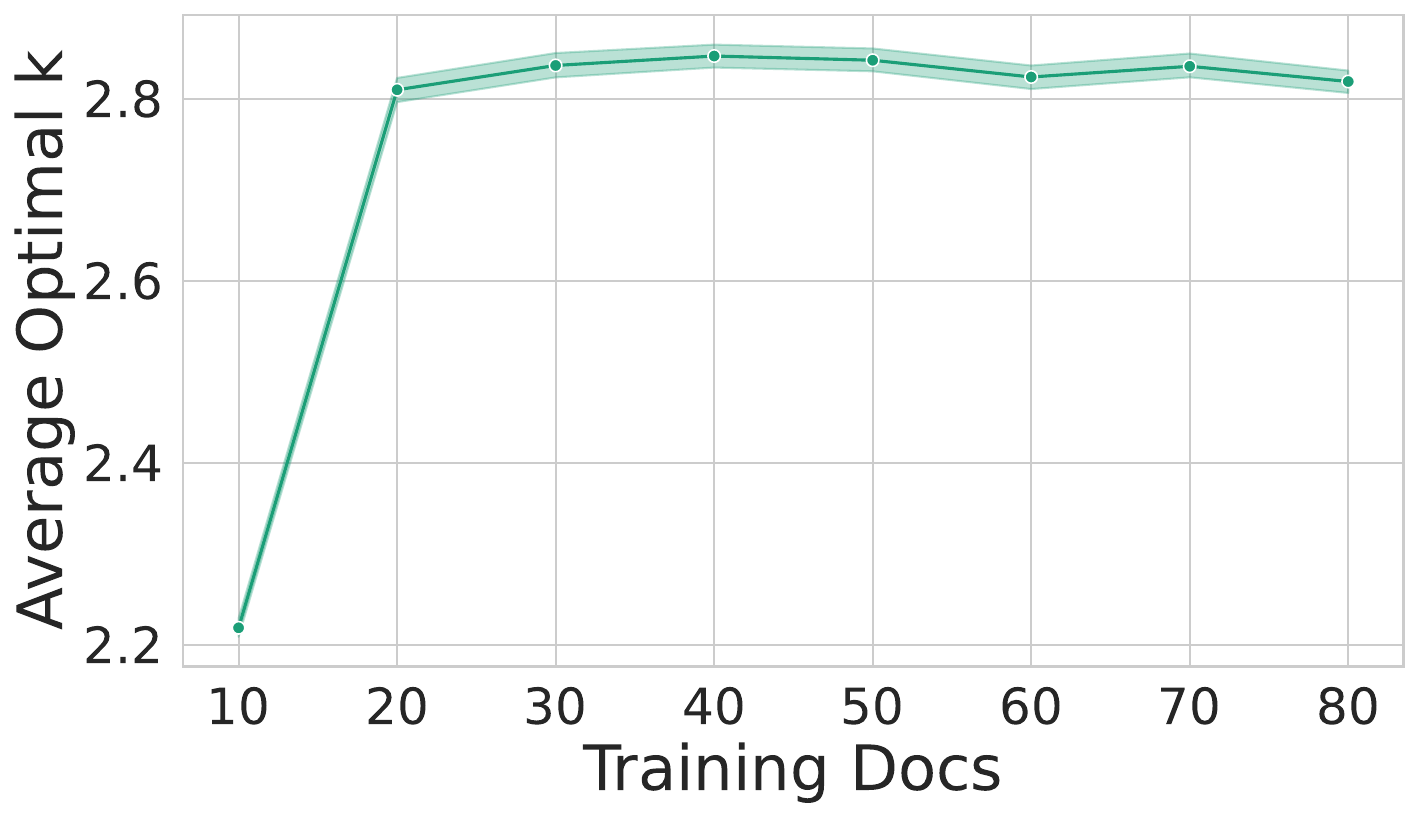}
  \caption{The mean optimal number of Gaussian components, $k$, plateaus after approximately 20 training documents, indicating diminishing returns from additional samples.}
  \label{fig:avg_k}
\end{figure}

Figure~\ref{fig:avg_k} shows the mean number of GMM components after fitting each model on a fixed number of training documents, choosing the $k$ value with the highest log-likelihood on the held-out validation data. The results suggest that even with relatively few documents for an author, e.g., 10, some authors have sufficiently distinct but regular stylistic variation that benefits from having multiple components. The number of optimal components stagnates as soon as we get to 20 training documents, indicating that a small number of documents is all that's needed to gain benefit from this modeling setup. 

Table~\ref{tab:k_distribution} shows the relative frequency of authors having a specific number of GMM components, $k$, in Experiment 1 (Section~\ref{sec:experiment}).

\begin{table}[htbp]
  \centering
  \begin{tabular}{cr}
    \hline
    \textbf{GMM Components} & \textbf{Percentage of Authors} \\
    \hline
    1 & 20.94 \\
    2 & \textbf{33.80} \\
    3 & 17.49 \\
    4 & 9.71 \\
    5 & 6.70 \\
    6 & 5.77 \\
    7 & 5.60 \\
    \hline
  \end{tabular}
  \caption{We find that roughly 61\% of authors have optimal $k$ values between 2-4. Additionally, only about 20.94\% of authors have the centroid as their best model, indicating that an overwhelming majority of authors benefit from more complex modeling. Interestingly, 18\% of authors have 5+ components, indicating a much higher idiolectic complexity than the average author.}
  \label{tab:k_distribution}
\end{table}

\subsection{Does the type of document style influence the optimal $k$?}

\begin{figure}[htbp]
  \centering
  \includegraphics[width=0.47\textwidth]{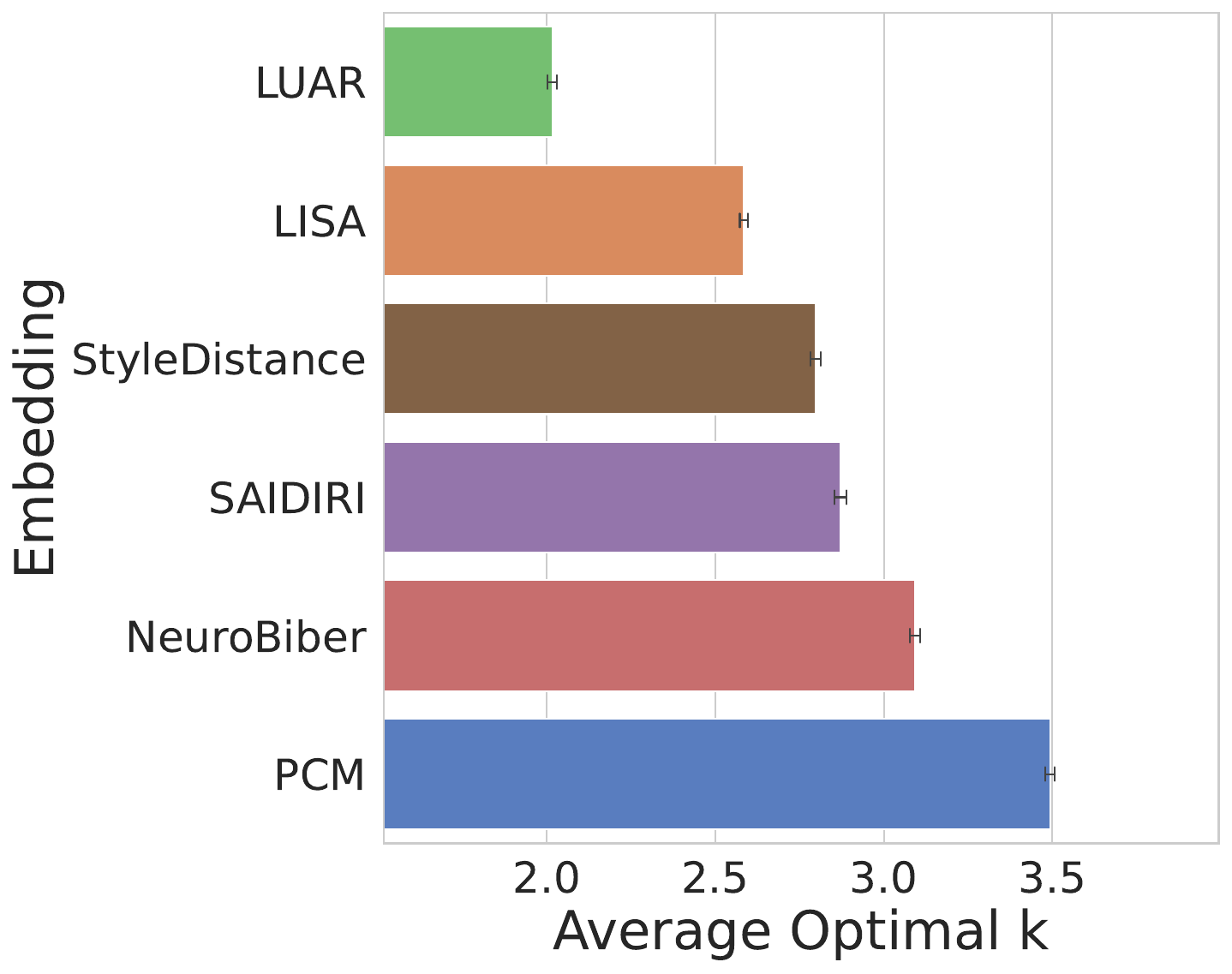}
  \caption{There does not seem to be a general trend with respect to the strength of the embedding and the optimal $k$ value. In fact, the weakest embedding (NeuroBiber) has a higher average number of components than the strongest embedding (SADIRI). }
  \label{fig:k_embeddings}
\end{figure}

Here, we examine the distribution of optimal $k$ values across the different types of document styles, shown in Figure~\ref{fig:k_embeddings}. We can see that the PCM embedding leverages the modeling complexity the most. However, there does not seem to be a correlation with the results in Figure~\ref{fig:experiment_comparison}, as the SADIRI based embedding had the highest performance boost from GMM modeling but not the most complex distributions. Overall, there is not a clear relationship between the style representation performance and the average optimized $k$.  

\section{Orthogonalization Analysis}
\label{app:orthogonalization}

\subsection{Stylistic Overlap Amongst Embeddings}
\label{app:stylistic-overlap}

\begin{figure}[htbp]
  \centering
  \includegraphics[width=0.47\textwidth]{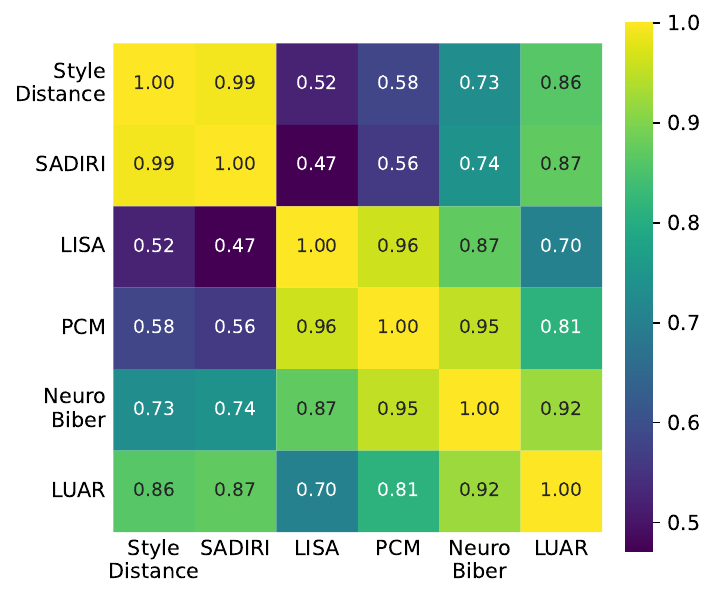}
  \caption{There is a clear overlap in the stylistic features that the six embeddings are capturing. }
  \label{fig:evr}
\end{figure}

We hypothesize that the discrepancy in performance across embeddings arises from differences in the stylistic information captured by each representation. In particular, we expect the SADIRI embedding to capture much of the stylistic variance represented in NeuroBiber, while additionally encoding a broader set of latent features. To test this, we measure the Expected Variation Ratio (EVR) across the embeddings. First, we perform Singular Value Decomposition (SVD) to bring all of the embeddings down to 96 dimensions. We then compute the EVR for each embedding, which represents the proportion of variance explained by each of the 96 latent dimensions obtained from SVD. By correlating these EVR profiles, we observe in Figure~\ref{fig:evr} that the principal components of these embeddings exhibit highly similar variance structures, suggesting that they encode overlapping stylistic information. At an aggregate level, the embeddings exhibit consistently high correlations, indicating the presence of a shared set of stylistic signals distributed across their latent spaces. 

The heatmap additionally reveals several clustering patterns among the embeddings. StyleDistance and SADIRI exhibit nearly identical EVR structures, while LISA and PCM form a second highly correlated pair. NeuroBiber and LUAR maintain strong correlations across nearly all embeddings, suggesting that they may capture a more generalized stylistic representation. The near-perfect correlation between StyleDistance and SADIRI is especially interesting because StyleDistance is explicitly trained to be content-independent. The shared variance structure suggests that the dominant latent directions captured by the SADIRI embedding are not based on semantics or topic, but are strongly organized around stylistic variation. Further, these results suggest that the performance improvements observed in Figure~\ref{fig:experiment_comparison} are likely driven by stylistic structure.

\subsection{Performance Impact From Orthogonalization}

\begin{figure}[htbp]
  \centering
  \includegraphics[width=0.47\textwidth]{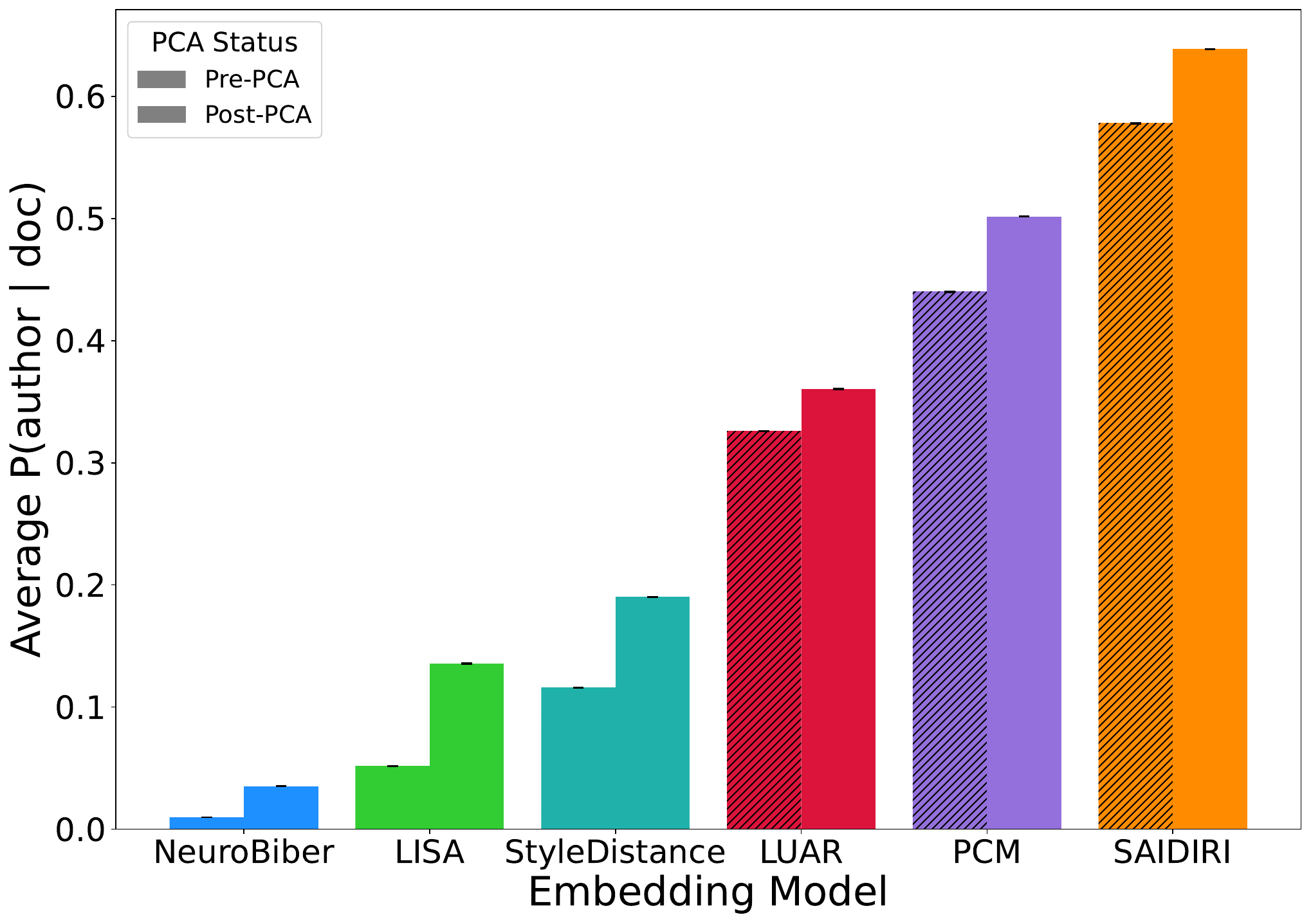}
  \caption{There is a large boost to performance for all embeddings after using orthogonalization in our modeling setup. This indicates that the step is crucial for our diagonal covariance setup.}
  \label{fig:pca_performance}
\end{figure}

Although orthogonalization is relatively cheap in terms of time complexity, one might question the impact on performance. In this experiment, we analyze the performance of the mixed author model setup with GMMs trained with a diagonal covariance matrix with and without orthogonalization. Indeed, the experiments without orthogonalization were quicker, but from Figure~\ref{fig:pca_performance}, we can clearly see a strong jump in performance for every embedding. LISA and StyleDistance observe the most performance gain, which is surprising because they have a much lower dimensionality than PCM and SADIRI. These results suggest that the raw embedding spaces violate the independence assumptions imposed by diagonal covariance GMMs. Orthogonalization alleviates this by decorrelating the latent dimensions, which satisfies a key assumption of diagonal covariance matrix GMM modeling. 

\subsection{PCA Time Complexity Analysis}

Now that we've established that orthogonalization gives us a major performance boost, we look to formalize the time complexity benefit from using a diagonal covariance matrix instead of a full one.  

\noindent \textbf{Setup.}
Let $n$ be the number of samples, $d$ the embedding dimension, $k$ the number of mixture components, and $T$ the number of EM iterations.

\noindent \textbf{Full covariance matrix.}
Per EM iteration, evaluating Mahalanobis distances and updating full covariances yields
\[
\text{per-iter: } O(nkd^2 + kd^3),
\qquad
\text{total: } O\!\left(T\,(nkd^2 + kd^3)\right).
\]

\noindent \textbf{PCA + Diagonal Covariance.}
A one-time PCA cost for covariance formation and eigendecomposition:
\[
\text{one-time: } O(nd^2 + d^3).
\]
After orthogonalization (no projection, so $r=d$), diagonal-covariance EM steps require only elementwise operations:
\[
\text{per-iter: } O(nkd),
\qquad
\text{total: } O(nd^2 + d^3) + O\!\left(T\,nkd\right).
\]

\noindent \textbf{Evaluation in Practice}
We use scikit-learn's implementation of GMMs with a max iteration of 100.
In our setup, the most computationally expensive case is
$n=2000000$, $d=2560$, $k=7$

Under these worst-case parameters with $T=1$, we see a speedup of approximately 7x. In the worst case $T=100$, and the
full covariance is approximately $\mathbf{550}\times$ more work overall 
than PCA + diagonal covariance.

\subsection{Dimensionality Reduction}

\begin{figure}[htbp]
  \centering
  \includegraphics[width=0.48\textwidth]{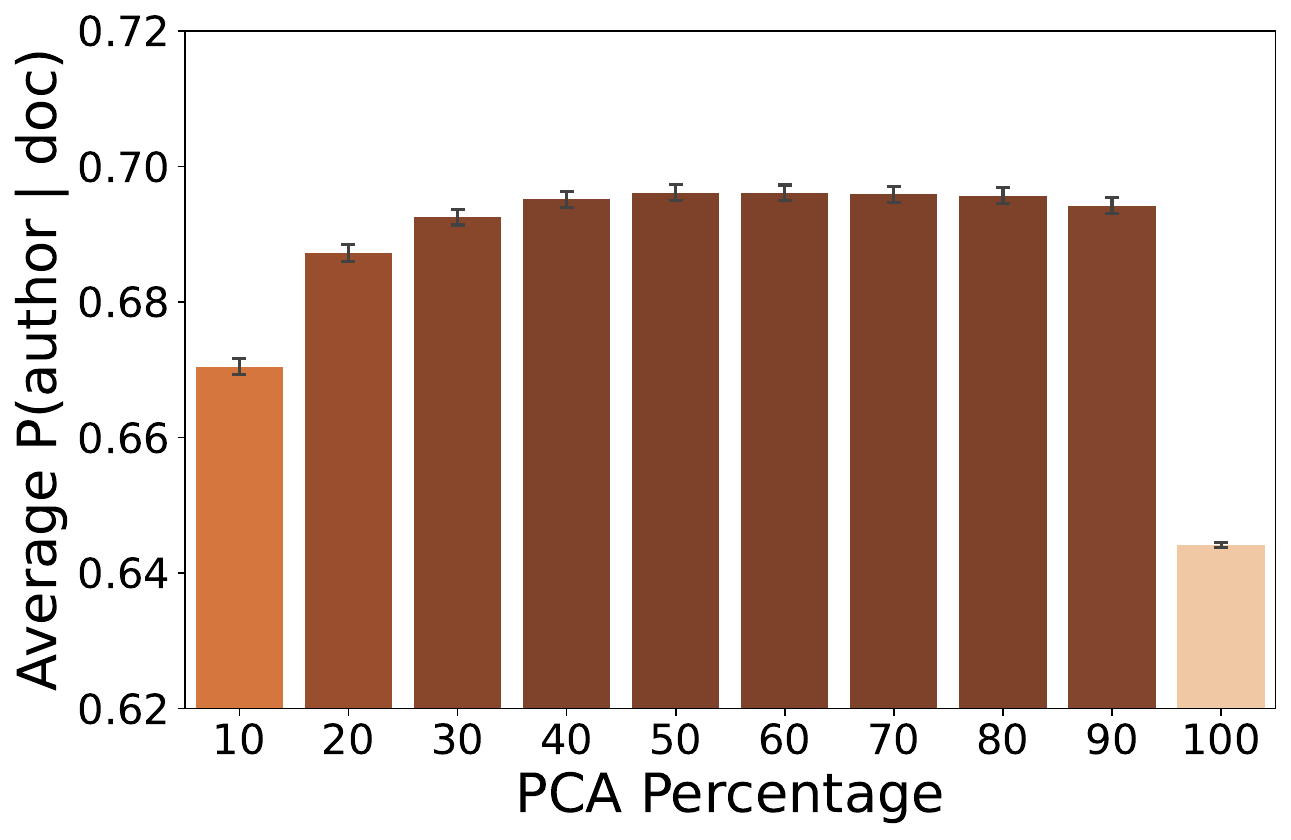}
  \caption{We need not perform full orthogonalization to gain most of the benefits. This figure shows that we can use 50\% of the PCA reduced embedding and get the best performance for a much lower computational cost.}
  \label{fig:pca_varying}
\end{figure}

We have shown that orthogonalization provides strong performance gains in Figure~\ref{fig:pca_performance} and negligible overhead when compared to using a full covariance matrix. However, we can also leverage the dimensionality reduction aspect of PCA and fit our models on a reduced embedding. In Figure~\ref{fig:pca_varying}, we run PCA on a varying percentage of the SADIRI embedding on 10\% of authors. Here, we observe an interesting phenomenon where performance is almost bell shaped where 50-60\% has the highest performance. Surprisingly, using the full embedding orthogonalized performs the worst, which indicates that some of the features are introducing noise. The result shows that we could speed up the modeling process substantially and get \textit{better} results.

Note that the raw consistency values are higher than those of the 40K author setup since this result is computed using only a subset of 4K authors due to performance considerations.

\section{Are authors consistent across genre changes?}
\label{app:wiki_genre}

Another interesting fluctuation in writing style occurs when authors write in different genres. Linguistic literature shows us that authors are often consistent across multiple genres \citep{mojedano2024epistemic}. Another study has shown that an author's idiolect tends to have low intra-author variation when writing about different topics \citep{litvinova2018assessing}. We look to evaluate these findings by observing authors who write in Wikipedia article pages and talk pages.

\noindent \textbf{Setup}
First, we filter our dataset to include only posts from authors who have at least 10 posts in both article and talk pages. Next, we compute the quartiles with respect to consistency for article pages. We evaluate these sets of authors together to determine how their performance changes when evaluated in the talk page setting. For context, article pages are the encyclopedia articles that provide factual information regarding subjects. Talk pages serve as a forum for improving article pages, so they are inherently much less formal. 

\noindent \textbf{Results}
Surprisingly, we see that the most consistent authors in article pages are the least consistent in talk pages. On the other hand, the least consistent authors in formal settings seem to be the most consistent in the informal setting. The key takeaway here is that authors who write stylistically consistently in one setting are not guaranteed to do the same in another setting. Although there are multiple studies supporting the claim that the author's idiolect remains consistent across genres, these studies only look at fewer than 100 authors. Our analysis for Wikipedia captures posts from over 2000 authors; thus, it could be plausible that these papers did not have a large enough sample size to evaluate their claims. 

\begin{figure}[htbp]
  \centering
  \includegraphics[width=0.47\textwidth]{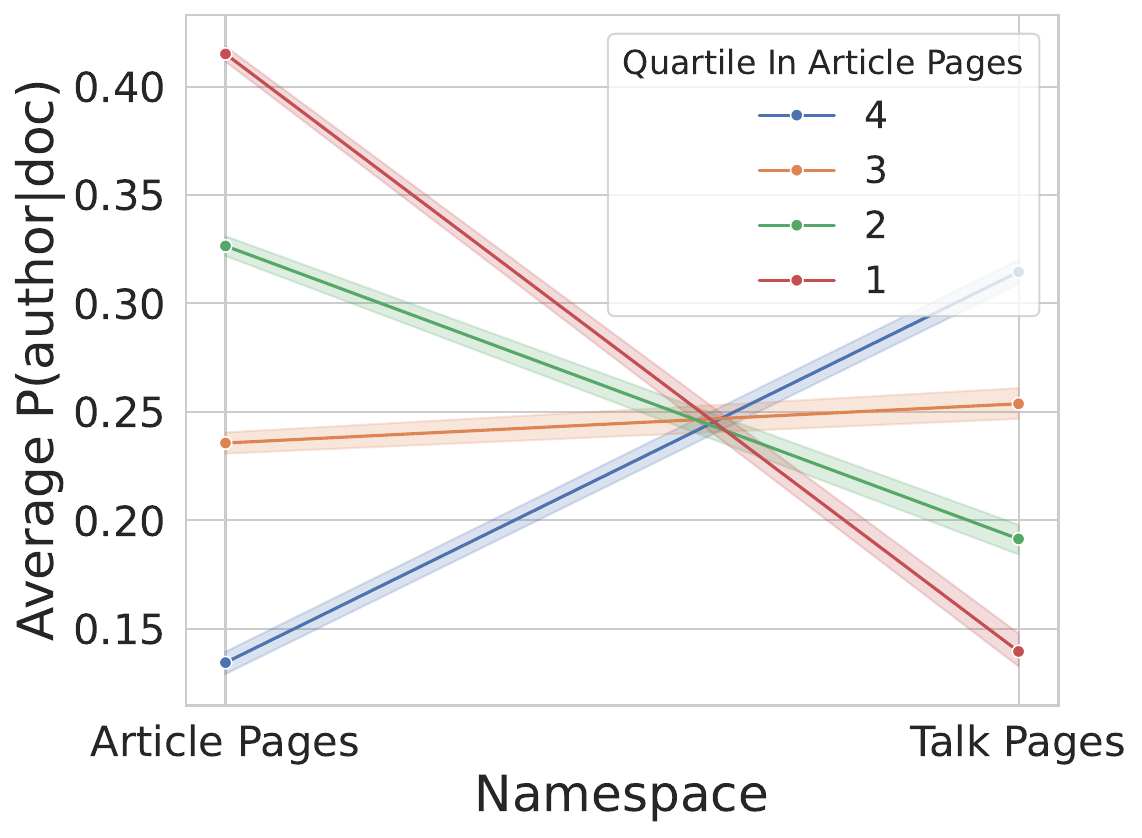}
  \caption{Authors who are consistent in one genre are not guaranteed to be consistent in another. This Wikipedia analysis shows us that stratified authors who are consistent in a formal setting actually are the least consistent authors in a more informal setting.}
  \label{fig:wiki_formality}
\end{figure}

\section{Additional Details on Temporal Analysis}
\label{app:temporal-analysis}

Here, we fully specify how models were constructed for testing the three hypotheses in the temporal analysis in Section~\ref{sec:temporal}. The documents for each author are partitioned into distinct windows.

To evaluate temporal generalization, we partition each author's chronologically ordered documents into non-overlapping, author-relative time windows. For an author whose earliest document occurs at time $t_0$, window $i$ is the interval $[t_0 + iW,, t_0 + (i+1)W)$, where $W$ is a fixed calendar span ($W \in {0.5, 1.5, 2.5}$ years). The number of documents per window varies by author; only authors with at least $m$ documents in a window are included ($m \in {15, 20, 25}$), and an author is filtered out if they do not appear in at least 3 windows. Further, a window is retained only if at least 10\% of the \textbf{remaining} author set exists in it.  Since we pull from 3 sources, we need at least 3k authors in each window to make the window valid.

Within each window, documents are partitioned into train, dev, and test splits with proportions 60\%, 20\%, and 20\%. The split is random but deterministic for each author–window pair, so the same documents always fall into the same split. A separate GMM is trained on the training portion of each window's data, with the number of components $k$ chosen via the dev set. Performance is computed by evaluating each model trained on window $i$ on the test documents from every window $j$ (including $j = i$), where mean log-likelihood is the metric. 

\begin{figure}[tbh]
    \centering
    \includegraphics[width=0.98\linewidth]{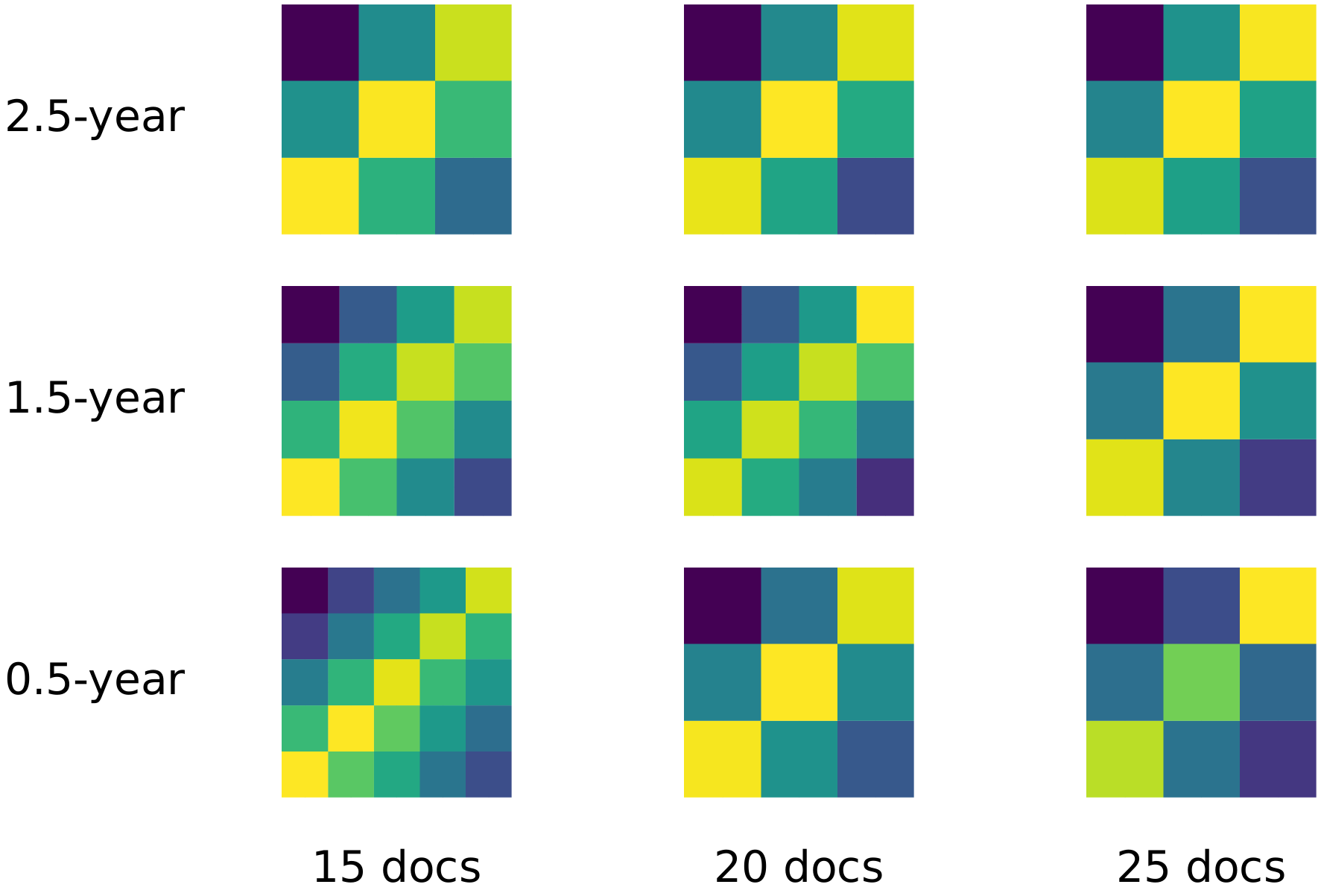}
    \caption{The grid tabulates different hyperparameter configurations for the time heatmap found in Figure~\ref{fig:time_heatmap}. We can see that the general trend of the diagonal performing the best holds up regardless of the time range and minimum number of documents.}
    \label{fig:time_grid}
\end{figure}

Figure~\ref{fig:time_heatmap} shows one of these configurations where each window is half a year, and we require an author to have at least 15 documents in that range to be considered in the experiment. However, the phenomenon we observe in Figure~\ref{fig:time_heatmap} is robust to other hyperparameter choices, as shown in Figure~\ref{fig:time_grid}. Here we tabulate all of our configurations for different time ranges and minimum number of documents to show that the same overall trend exists where the diagonal is the most performant, reinforcing the claim that H1 is not founded. 

\begin{figure}[tbh]
    \centering
    \includegraphics[width=0.98\linewidth]{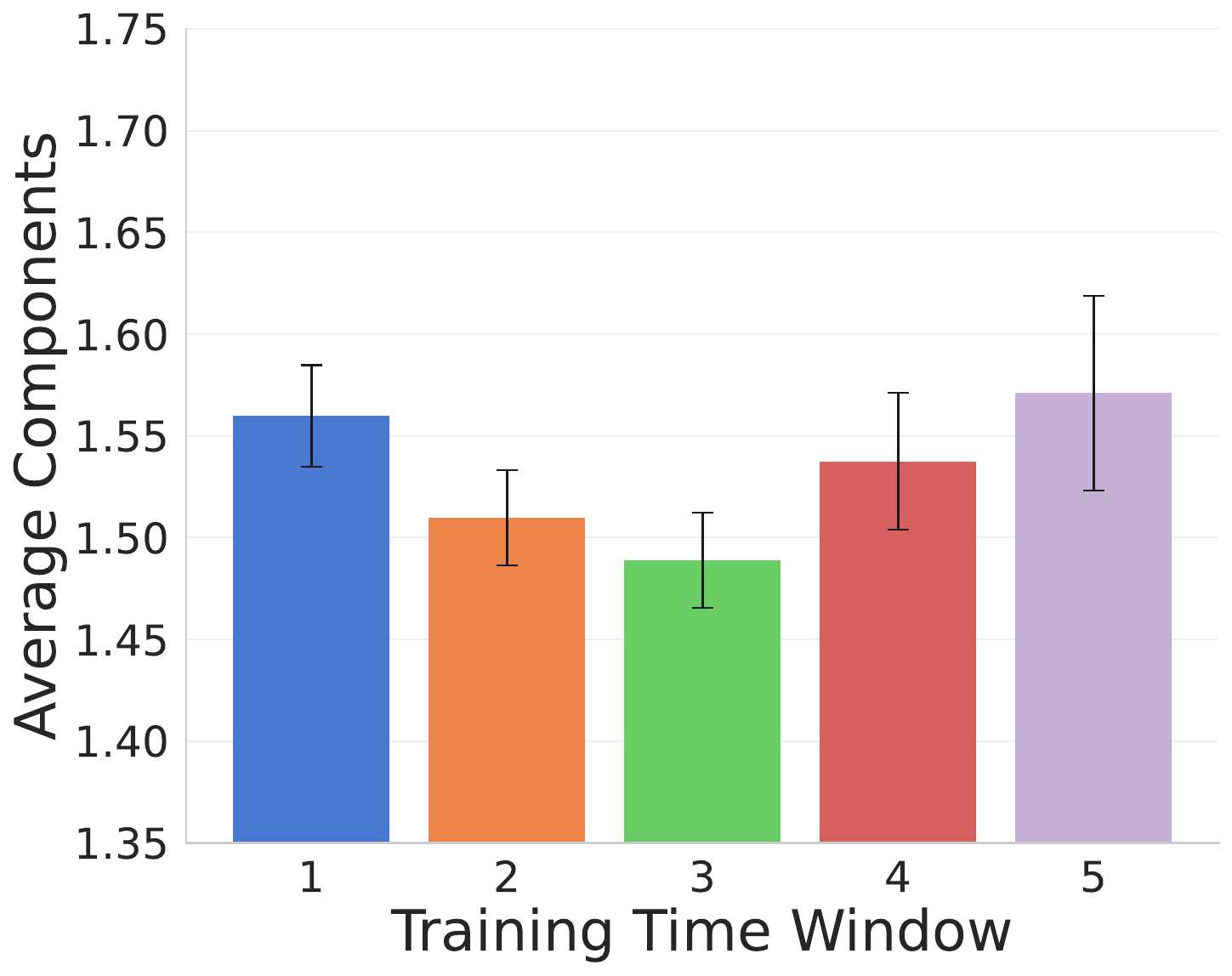}
    \caption{The mean number of components per GMM across the different temporal windows used in the experiment in Section~\ref{sec:temporal}; cf. Figure~\ref{fig:time_heatmap}. Note that the same authors are present in all windows, so the changes reflect a shift in global behavior. }
    \label{fig:temporal-components}
\end{figure}

To assess whether authors' styles become more or less complex over time, we calculate the mean number of components used by the same set of authors in each time window, shown in 
Figure~\ref{fig:temporal-components}. The number of components can be thought of as a loose proxy for stylistic diversity, where authors with minimal variation are more likely to be modeled with a single component, whereas those with multiple distinct styles are more likely to have multiple components. Here, we see that authors have negligible variation over time. This indicates that H2 is not supported. 

\begin{figure}[tbh]
    \centering
    \includegraphics[width=0.98\linewidth]{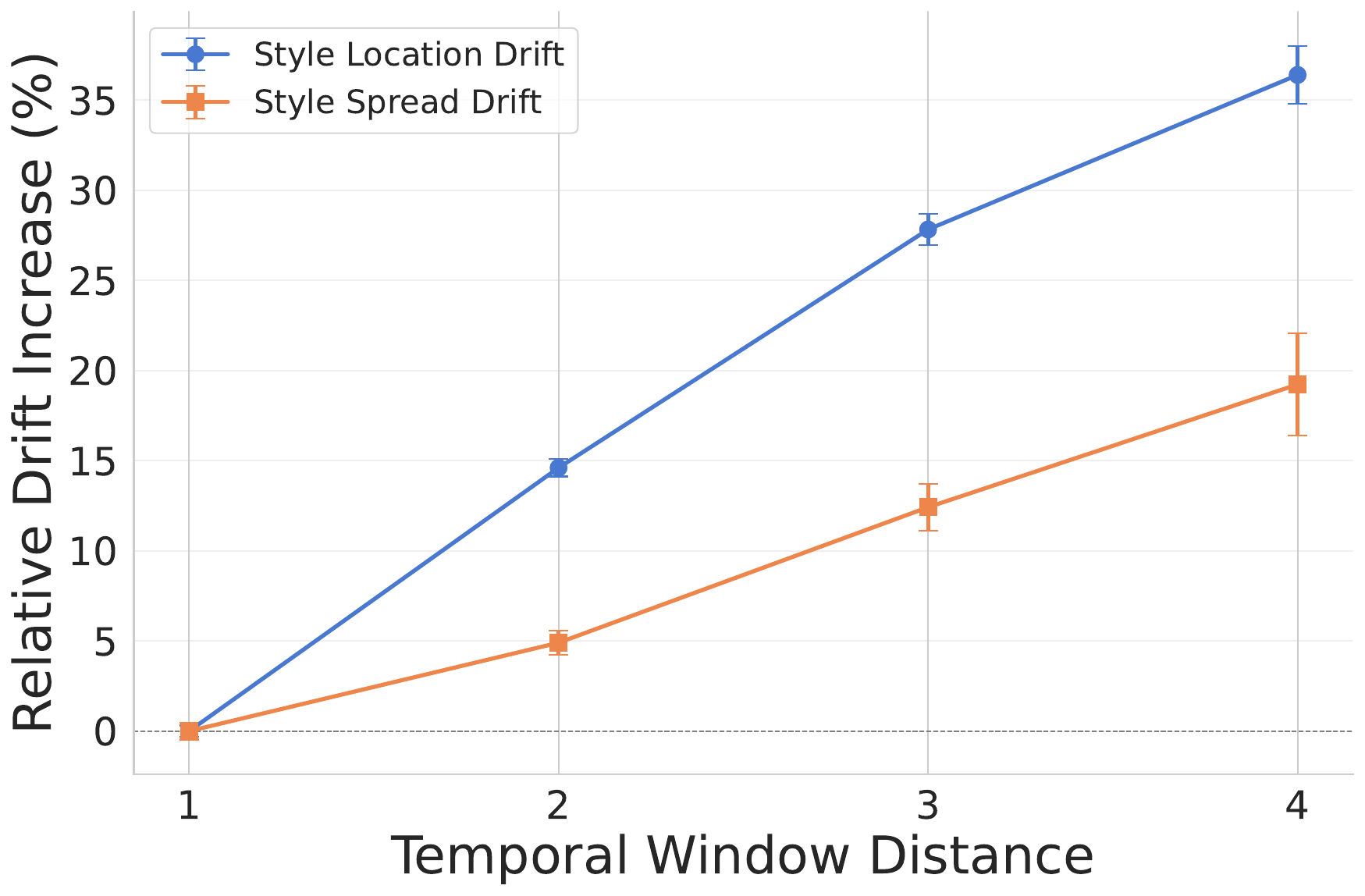}
    \caption{Relative increase in location drift ($D_\mu$) and spread drift ($D_\sigma$) as a function of temporal lag $|i-j|$. Location drift grows substantially faster than spread drift as lag increases, indicating that temporal degradation is driven primarily by stylistic displacement rather than inconsistency. Note we use the same configurations as Figure~\ref{fig:time_heatmap}.}
    \label{fig:temporal-lag}
\end{figure}

To investigate H3, we quantify how much GMM \emph{location} and \emph{spread} diverge across calendar windows as temporal distance grows. We take the learned models that were learned from the configurations in Figure~\ref{fig:time_heatmap}. For each pair of temporal windows $(i,j)$, we compute a location drift score,
\[
D_\mu(i,j)=\left\lVert \boldsymbol{\mu}_i-\boldsymbol{\mu}_j \right\rVert_2,
\]
and a spread drift score,
\[
D_\sigma(i,j)=\left\lVert \log \boldsymbol{\sigma}_i^2-\log \boldsymbol{\sigma}_j^2 \right\rVert_2.
\]
We then group these pairwise comparisons by temporal lag $|i-j|$. For example, lag~1 compares adjacent windows ($1\!\leftrightarrow\!2$, $2\!\leftrightarrow\!3$, etc), while lag~2 compares windows separated by one interval ($1\!\leftrightarrow\!3$, $2\!\leftrightarrow\!4$, etc). Finally, for each lag, we report the average drift as a percentage increase relative to lag~1 as shown in Figure~\ref{fig:temporal-lag}.

Location drift increases substantially faster with temporal distance than spread drift. Relative to lag~1, the mean location drift $D_\mu$ increases by $14.6\%$ at lag~2, $27.8\%$ at lag~3, and $36.4\%$ at lag~4. In contrast, the mean spread drift $D_\sigma$ increases more gradually, by only $4.9\%$, $12.4\%$, and $19.2\%$ at the corresponding lags. Together with the stable component counts in Figure~\ref{fig:temporal-components}, these results support H3: authors' stylistic representations primarily translate through embedding space over time, while overall mixture complexity remains approximately stable and changes in spread accumulate more slowly.

Importantly, $D_\sigma$ measures the magnitude between spread profiles of temporal windows rather than a strict increase in variance over time. When we instead compare each author's earliest and latest windows directly, $60.2\%$ of authors exhibit tighter distributions, while $39.8\%$ exhibit broader distributions. Overall, this produces only a modest net tightening, but there is no general phenomenon that a majority of authors follow. 

\section{Conversation Style Shifts}
\label{app:conversation}

\begin{figure}[htbp]
    \centering
    \includegraphics[width=0.47\textwidth]{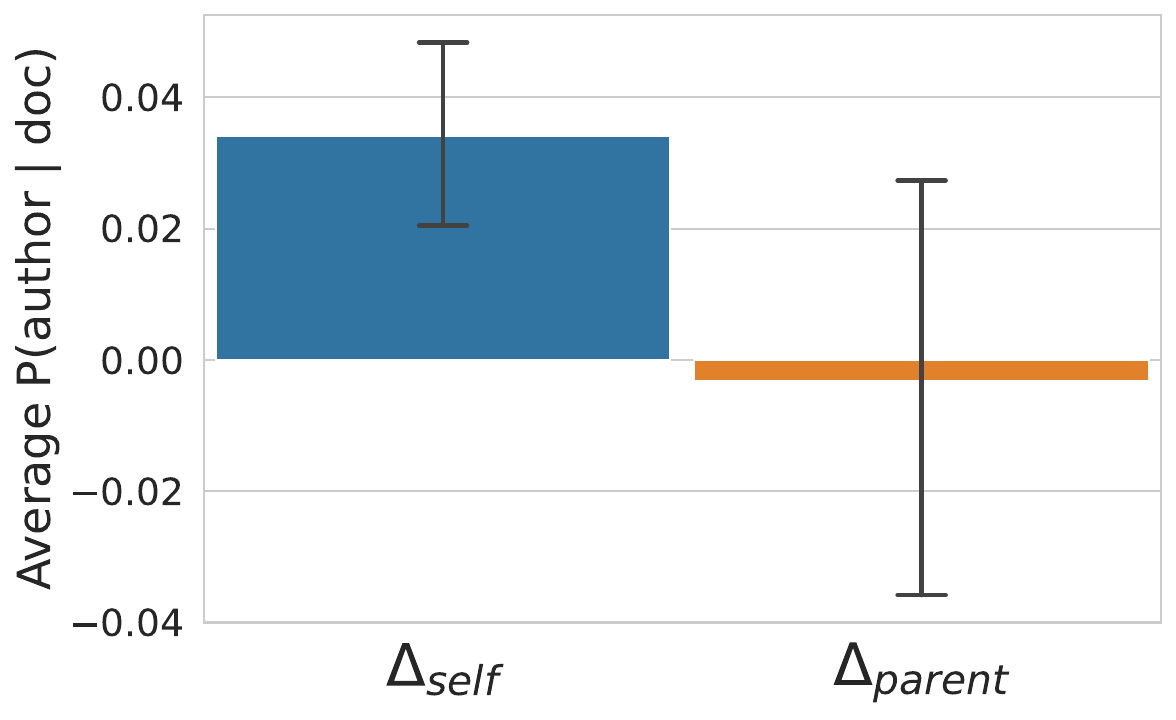}
    \caption{Average $\Delta_{\mathrm{self}}$ and $\Delta_{\mathrm{parent}}$ values for GMANE replies. Replies become slightly more idiolectically rich for the reply author, while showing no meaningful accommodation. }
    \label{fig:gmane}
\end{figure}

Figure~\ref{fig:gmane} shows the result in the conversation style shift experiment. We observe that replies are slightly richer in idiolect for the reply author. Additionally, reply authors do not seem to shift their style toward the parent author's idiolect distribution. 

\section{Additional Training Details}
\label{app:training_details}

The RoBERTa model was trained to predict the specific author from a document. Documents are clipped at 512 tokens. The model is trained for 3 epochs, and we use a downsampled 100K development set to choose the best-performing model across the epochs. We use \texttt{roberta-base} as the base model, a batch size of 32, and the AdamW optimizer with a learning rate of 2e-5, weight decay of 0.01, and 10 warm-up steps. We did not perform extensive hyperparameter tuning for the RoBERTa baseline; instead, we used a standard configuration to provide a strong and reproducible reference point. We observed similar poor performance when using DistilRoBERTa, indicating that the particular model is likely not causing performance degradation.

For all of our experiments, we trained and evaluated on NVIDIA RTX A6000 GPUs. For the nearest neighbor and centroid approaches, roughly 4 GPU hours on one A6000 were sufficient for training and evaluating our models. In the GMM case, we needed up to 24 GPU hours for training and evaluating on larger embeddings, like the SADIRI embedding. The RoBERTa experiment required roughly 48 GPU hours for training and evaluation. 

\subsection{Runtime Details}

For runtime comparison across models, Table~\ref{tab:runtime_complexity} shows the worst-case time complexity across the fitting and evaluation steps of the experiments, extending the PCA analysis in Appendix~\ref{app:orthogonalization}. Let $A$ be the number of authors, $S$ the number of test documents per author, $k$ the number of components per author, $d$ the embedding dimension, and $T$ the number of EM iterations per author.

\begin{table}[htbp]
\centering
\begin{tabular}{lcc}
\hline
\textbf{Model} & \textbf{Fitting} & \textbf{Evaluating} \\
\hline
GMM & $O(AkdT)$ & $O(A^2Skd)$ \\
Centroid & $O(AdT)$ & $O(A^2Sd)$ \\
Nearest Neighbor & $O(Ad)$ & $O(A^2Sd)$ \\
\hline
\end{tabular}
\caption{Worst-case fitting and evaluation complexity for the author models.}
\label{tab:runtime_complexity}
\end{table}

For training, we set $T$ to be capped at 100. Additionally, from Table~\ref{tab:k_distribution}, we know that the average $k$ value is approximately 2.64. Thus, in the worst case, nearest-neighbor training is 100 times faster than the centroid model and 264 times faster than the GMM model. Evaluation runtime is more comparable, where the GMM model is roughly 2.64 times slower because of the varying component size. These runtime complexities are informative because experiments varied in which CPU was used and may be affected by other system-dependent bottlenecks, such as I/O, at the time of the experiment.

\section{Performance Metrics}
\label{app:performance_metrics}

Below we provide two tables that more cleanly represent the values for the performance metrics in Figure~\ref{fig:experiment_comparison}. 

\begin{table}[H]
\centering
\small

\resizebox{\columnwidth}{!}{%
\begin{tabular}{@{}lrrrr@{}}
\toprule
\textbf{Representation}
& \textbf{NN}
& \textbf{Centroid}
& \textbf{GMM}
& \boldmath$\Delta$ \\
\midrule
NeuroBiber    & 0.15 & 0.07 & 0.08 & +0.01 \\
LISA          & 0.17 & 0.23 & 0.26 & +0.03 \\
StyleDistance & 0.23 & 0.29 & 0.34 & +0.05 \\
LUAR          & 0.47 & 0.55 & 0.57 & +0.02 \\
PCM           & 0.64 & 0.65 & 0.69 & +0.04 \\
SADIRI        & 0.71 & 0.76 & 0.78 & +0.02 \\
\bottomrule
\end{tabular}%
}

\caption{
Success@8 values corresponding to
Figure~\ref{fig:experiment_comparison}.
NN denotes the nearest-neighbor model, and $\Delta$ represents the absolute improvement from using the GMM model versus the Centroid model.
}
\label{tab:model_performance_success}
\end{table}

\vspace{-4pt}

\begin{table}[H]
\centering
\small

\resizebox{\columnwidth}{!}{%
\begin{tabular}{@{}lrrrr@{}}
\toprule
\textbf{Representation}
& \textbf{NN}
& \textbf{Centroid}
& \textbf{GMM}
& \boldmath$\Delta$ \\
\midrule
NeuroBiber    & 0.00 & 0.03 & 0.04 & +0.01 \\
LISA          & 0.00 & 0.11 & 0.14 & +0.03 \\
StyleDistance & 0.00 & 0.15 & 0.19 & +0.04 \\
LUAR          & 0.00 & 0.34 & 0.36 & +0.02 \\
PCM           & 0.00 & 0.44 & 0.50 & +0.06 \\
SADIRI        & 0.00 & 0.61 & 0.64 & +0.03 \\
\bottomrule
\end{tabular}%
}

\caption{
Average $p(a\mid d)$ values corresponding to
Figure~\ref{fig:experiment_comparison}.
NN denotes the nearest-neighbor model, and $\Delta$ represents the absolute improvement from using the GMM model versus the Centroid model.
}
\label{tab:model_performance_prob}
\end{table}

\section{AI Assistance}
ChatGPT was used for assistance with coding, specifically in evaluation optimizations on GPUs. In addition, we used AI assistance for grammatical corrections. 

\end{document}